\documentclass[10pt,twocolumn,letterpaper]{article}

\newcommand{\model}{MP-GUI\xspace}

\usepackage{tcolorbox}
\tcbuselibrary{breakable}

\usepackage[pagenumbers]{cvpr} 
\usepackage{mathtools}

\usepackage{float}
\usepackage{colortbl}
\definecolor{myrose}{rgb}{0.99, 0.91, 0.95}
\definecolor{mygreen}{rgb}{0, 0.6, 0}
\usepackage{makecell}
\usepackage{multirow}

\definecolor{cvprblue}{rgb}{0.21,0.49,0.74}
\usepackage[pagebackref,breaklinks,colorlinks,allcolors=cvprblue]{hyperref}

\title{MP-GUI: Modality Perception with MLLMs for GUI Understanding }

\author{
Ziwei Wang$^{1 \dagger}$, Weizhi Chen$^{1 \dagger}$, Leyang Yang$^{1}$, Sheng Zhou$^{2*}$, Shengchu Zhao$^{3}$, Hanbei Zhan$^{1}$,\\ 
Jiongchao Jin$^{3}$, Liangcheng Li$^{1}$, Zirui Shao$^{1}$, Jiajun Bu$^{1}$\\
$^1$College of Computer Science and Technology, Zhejiang University, China\\
$^2$Zhejiang Key Laboratory of Accessible Perception and Intelligent \\Systems, Zhejiang University, China\\
$^3$Ant Group\\
\tt\small \{wangziwei98, zhousheng\_zju, chenweizhi, yangleyang, beiii7533\}@zju.edu.cn\\
\tt\small \{shaozirui,liangcheng\_li, bjj\}@zju.edu.cn
\tt\small \{shengchu.sc,jinjiongchao.jjc\}@antgroup.com
}

\usepackage{pdfpages}

\begin{document}
\maketitle
\renewcommand\thefootnote{}
\footnotetext{$^{*}$Corresponding author. $^{\dagger}$Equal Contribution.}
\renewcommand\thefootnote{\arabic{footnote}}

\begin{abstract}
Graphical user interface (GUI) has become integral to modern society, making it crucial to be understood for human-centric systems. However, unlike natural images or documents, GUIs comprise artificially designed graphical elements arranged to convey specific semantic meanings. Current multi-modal large language models (MLLMs) already proficient in processing graphical and textual components suffer from hurdles in GUI understanding due to the lack of explicit spatial structure modeling. Moreover, obtaining high-quality spatial structure data is challenging due to privacy issues and noisy environments. To address these challenges, we present \model, a specially designed MLLM for GUI understanding. \model features three precisely specialized perceivers to extract graphical, textual, and spatial modalities from the screen as GUI-tailored visual clues, with spatial structure refinement strategy and adaptively combined via a fusion gate to meet the specific preferences of different GUI understanding tasks. To cope with the scarcity of training data, we also introduce a pipeline for automatically data collecting. Extensive experiments demonstrate that \model achieves impressive results on various GUI understanding tasks with limited data. Our codes and datasets are publicly available at \textcolor{red}{\textit{\nolinkurl{https://github.com/BigTaige/MP-GUI}}}.

\end{abstract}

\section{Introduction} \label{sec:introduction}
Graphical user interface (GUI) is an important medium of human-computer interaction. 
Understanding GUI is critical in many human-centric interactions such as accessibility~\cite{edwards1995access}, automated agent systems~\cite{derksen2011agent,seeclick,lin2024showui,lu2024omniparser} and app testing systems~\cite{linares2017continuous}. 
The early research on GUI understanding mainly contributes to simple deep models ~\cite{li2021screen2vec,zhao2021guigan,bai2021uibert,pix2struct} to accomplish specific tasks. 
Recently, the rapid development of MLLMs~\cite{Qwen2VL,chen2024far,yao2024minicpm,Llama3} has achieved tremendous success in various visual understanding scenarios ~\cite{tong2024eyes,zhang2024differential,luo2024layoutllm,hu2024mplug,li2024chemvlm,you2023ferret} (\textit{e.g.}, natural image and documents) and shown great potential to GUI recently~\cite{you2025ferret,seeclick}.

\begin{figure}
  \centering
  \includegraphics[width=0.95\linewidth]{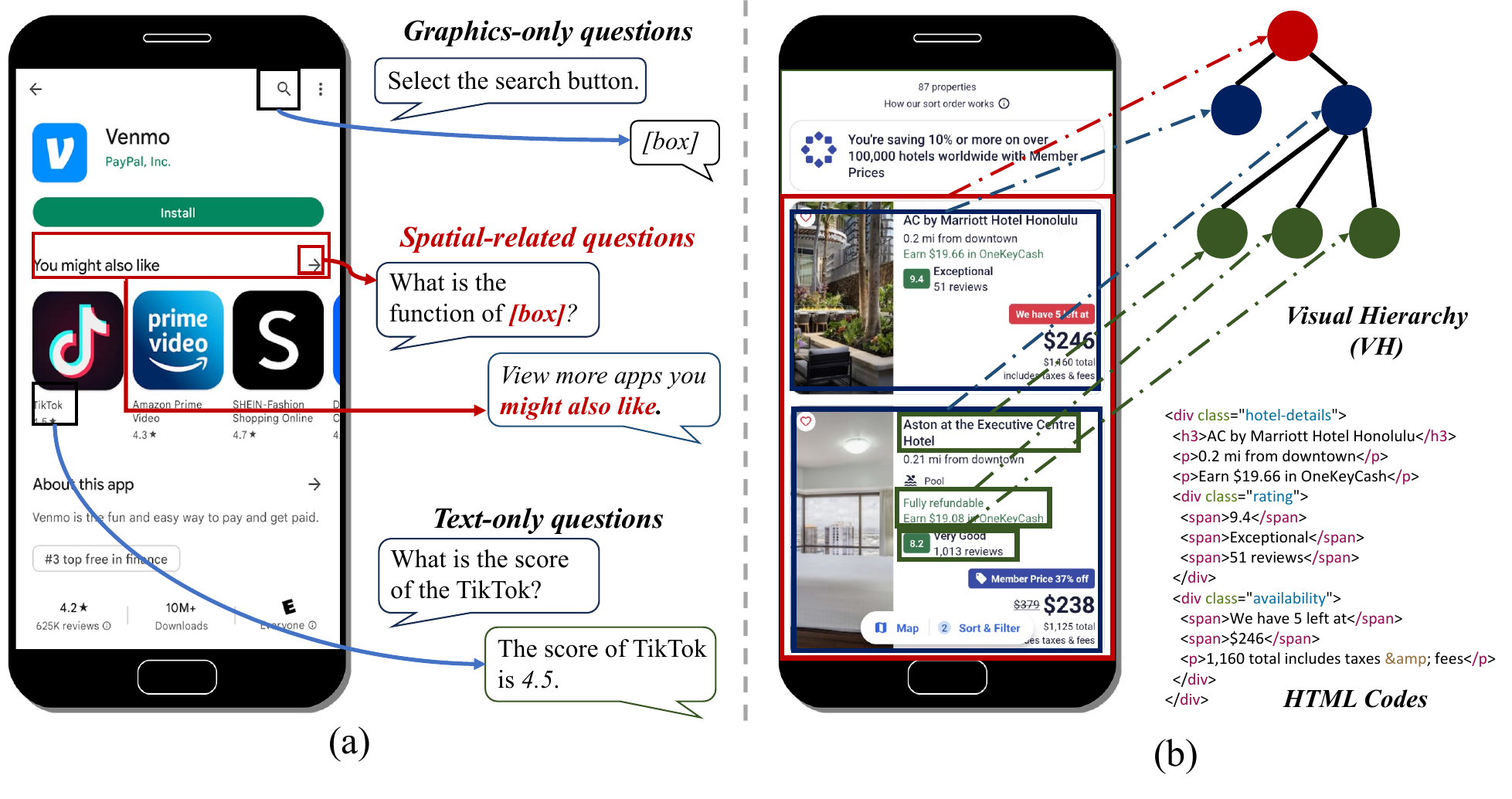}
  \caption{(a) Graphical, textual and spatial oriented tasks on GUI understanding. (b) Some classic GUI spatial structure forms.}
  \label{fig:intro}
\end{figure}

Although a few efforts have been made to utilize MLLMs for learning GUI knowledge\cite{you2025ferret,ma2024coco,bai2021uibert}, they typically treat the GUI screens as a natural images problem. 
Unlike objects in unstructured natural images that lack a pre-defined functional or semantic organization, GUI elements (\textit{e.g.}, icons/widgets and texts) are artificially designed and spatially arranged to convey specific semantic meanings, which are crucial for learning GUI knowledge.
For example, in \cref{fig:intro}-(a), answering the question {\tt "What is the function of the icon in the box?"} highly depends on the spatial relationship between the {\tt "you might also like"} text and {\tt "right arrow"} button that are semantically related. Otherwise, the {\tt "right arrow"} button will only convey the fundamental meaning of {\tt "going right"} and result in a wrong answer.
However, existing MLLMs for GUI understanding~\cite{spotlight,you2025ferret,seeclick} only rely on the original vision backbone\cite{vit,liu2021swin} to provide global screen visual clues and combine instruction fine-tuning to learn GUI knowledge.
As a result, despite their advanced capabilities in graphical and textual recognition~\cite{yao2024minicpm,chen2024far}, existing works still suffer 
from inaccurate feature representation after fusion and misunderstanding the GUI structure.

Meanwhile, accessing the screen spatial structure data is quite challenging in practice.
Some previous deep models have utilized manual and programmatic exploration~\cite{deka2017rico, rico_semantic} to obtain spatial structure information such as View Hierarchy (VH) and HTML~\cite{nakano2021webgpt} (\cref{fig:intro}-(b)).
However, they are often noisy~\cite{ross2018examining} and inconsistent with screenshots~\cite{li2020mapping}, failing to meet the demands of purely visual understanding.

To tackle the aforementioned challenges, we present \model, a dual visual-clues MLLM for visual GUI understanding. \model extracts graphical, textual, and spatial modality signals from the screen using three perceivers. This design reinforces the model's capability to grasp both visual and textual modalities by integrating spatial information, improving overall screen perception. Notably, each type of GUI understanding task may have distinct screen content preferences~\cite{widgetcaption,screenQA,refexp}. Thus, \model includes a Fusion Gate that adaptively combines screen signals of different modalities, enabling task-oriented feature fusion. Unlike current end-to-end implicit training\cite{seeclick,lin2024showui}, we design a training recipe including a Multi-stage Training Strategy and matching training objectives, especially a novel Spatial Relationship Prediction task and a data synthesis pipeline, which explicitly guides the model to learn GUI knowledge. Extensive experiments on various GUI-related benchmarks showcase the robust performance of \model. Our main contributions can be summarized as follows:

\textbf{(\textit{i})} We propose \model, which provides GUI-tailored visual clues for LLM via three perceivers and a semantically guided Fusion Gate, endowing the MLLM with effective GUI perception and understanding capability.

\textbf{(\textit{ii})} We introduce a training recipe with multi-stage strategies and task-specific objectives, which include a novel Spatial Relationship Prediction task and a data synthesis pipeline, to enable stage-wise explicit training of each perceiver in GUI knowledge learning.

\textbf{(\textit{iii})} We collect various GUI-related downstream tasks and the experimental results demonstrate the effectiveness of our special designs in the GUI scenarios.

\section{Related Work} 
\label{sec:related_work}
\subsection{Multi-modal Large Language Models}
Recently, Large Language Models (LLMs)~\cite{achiam2023gpt,gpt4,cai2024internlm2,qwen2.5} achieve remarkable results in various text tasks. On this basis, for text and vision modalities, most Multi-modal Large Language Models (MLLMs)~\cite{chen2024far,yao2024minicpm,Qwen-VL,Qwen2VL,Llama3,cogagent}, incorporate a pre-trained vision backbone, such as CNNs~\cite{lecun1989backpropagation,krizhevsky2012imagenet} or ViT~\cite{Qwen-VL}, and utilize a bridge module, like MLP~\cite{llava_1_5,chen2024far} or an attention-based resampler~\cite{blip-2}, to connect the vision backbone with the LLM for providing visual clues from images. Thanks to the powerful generic capabilities, instruction-tuning MLLMs on domain-specific data can quickly learns domain knowledge~\cite{liu2024visual}. ChemVLM~\cite{li2024chemvlm} utilizes chemically relevant instructional data to facilitate the acquisition of multimodal chemical knowledge by foundational models. Med-MLLM~\cite{liu2023medical} and MLeVLM~\cite{xu2024mlevlm} harness medical multimodal data to enhance the foundational model's performance within medical contexts. SeeClick~\cite{seeclick} collects GUI-related data to instruction-tune the foundational MLLM~\cite{Qwen-VL} for improving the grounding capability. Furthermore, ShowUI~\cite{lin2024showui} introduces a Visual Token Selection approach to reduce computational costs. It also integrates various GUI tasks in a systematic manner, aiming to effectively guide the MLLM~\cite{Qwen2VL} in learning domain knowledge. Generally, for most research on solving domain-specific multi-modal tasks, the dense image features generated by the vision backbone are the sole vision clues provided to the LLM. Given the distinctiveness of GUI scenarios, we introduce a dual-visual-clues framework, which provides the LLM with additional GUI-tailored visual clues, thus enhancing the screen perception capabilities of the MLLM.

\subsection{GUI Understanding}
GUI understanding involves a full range of downstream tasks, such as screen grounding~\cite{bai2021uibert,seeclick}, widget captioning~\cite{widgetcaption}, screen question answering~\cite{screenai,chen2021websrc}, navigation~\cite{aitw,mind2web}, screen summarization~\cite{screen2words}, etc. Considering the huge difference between GUI and general visual scenarios, many GUI-specialized models have been proposed\cite{spotlight,pix2struct,screenai,you2025ferret,bai2021uibert}. Pix2Struct~\cite{pix2struct} uses a dynamic patching strategy to adaptively divide the UI screen based on its resolution, along with a pre-training task of HTML reconstruction to allow the model to learn the UI knowledge. ScreenAI~\cite{screenai} uses a LLM~\cite{anil2023palm} to synthesize challenging screen schema data and integrates a large amount of UI-related pre-training data to improve model's UI and Infographics understanding. Ferret-UI~\cite{you2025ferret} uses image segmentation strategies to ensure that the original aspect ratio of mobile screenshots are not destroyed, and collects extensive GUI-related training data to improve the understanding of the foundational MLLM~\cite{you2023ferret} in GUI scenarios. Existing methods blend GUI-related data and implicitly learn GUI knowledge through unified end-to-end training.  In contrast, our approach uses three GUI-tailored perceivers integrated with a multi-stage training strategy. This enables training each perceiver with specific data at different stages, clarifying GUI knowledge learning via explicit training.

\begin{figure*}[ht]
  \centering
  \includegraphics[width=0.92\linewidth]{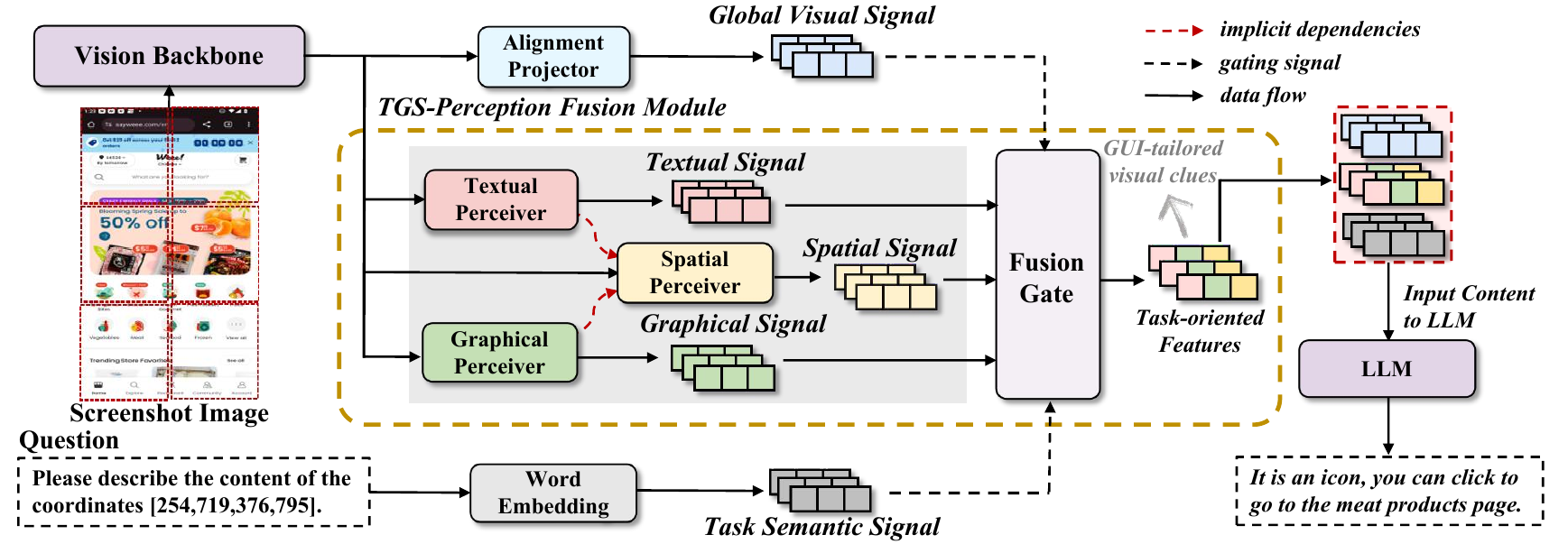}
  \caption{Overview of our \model. \model consists of three parts: (1) a vision backbone providing visual clues of the screenshot; (2) a TGS-Perception Fusion Module including three GUI-tailored perceivers for extracting specific GUI modality signals and a Fusion Gate for dynamically fusing these signals based on task semantics to produce GUI-tailored visual clues; and (3) an LLM generating results relying on screen visual clues, GUI-tailored visual clues, and task semantic signal.}
  \label{fig:model}
\end{figure*}


\section{Methodology} 
\label{sec:method}

\subsection{Model Architecture. }\label{sec:model_architecture}
As shown in \cref{fig:model}, \model is composed of three components: (1) a vision backbone; (2) a TGS-Perception Fusion Module (TGS-PFM) which includes three GUI-tailored perceivers and a Fusion Gate (FG) module; and (3) a LLM.

Given a screenshot image, we first utilize the Dynamic High-Resolution method~\cite{chen2024far} to divide the high-resolution image into several sub-images of 448$\times$448 pixels with optimal proportions. Let $\mathcal{I}\in \mathbb{R}^{N \times \mathcal{C}}$ denote the visual clues produced by vision backbone, where $N$ means the number of image tokens and $C$ is logits dimension. $\mathcal{Q}\in \mathbb{R}^{M \times \mathcal{D}}$ means the word embeddings of the question, where $M$ denotes the token length and $D$ is the dimension. Besides, visual clues $\mathcal{I}$ will be sent to an pre-trained Alignment Projector to align the dimention of LLM and get $\overline{\mathcal{I}}\in \mathbb{R}^{N \times \mathcal{D}}$.

\noindent \textbf{GUI-tailored Perceivers.}
Using the dense embeddings produced by vision backbone as image representation is a convenient paradigm to obtain visual clues~\cite{yao2024minicpm,cogagent,chen2024far}. However, the GUI information combined with visual clues is complex for several reasons: \textbf{\textit{(i)} screen images incorporate numerous multimodal elements and intricate details, which increase the difficulty of grounding for MLLMs; \textit{(ii)} there are high-level semantic associations between elements that need to be clearly identified, posing a challenge for MLLMs in understanding and being aware of the spatial relationships between GUI elements on the screen.} 

This complexity inhibits MLLMs from effectively handling GUI understanding tasks that rely solely on global visual clues. Thus, we propose the TGS-PFM, where three GUI-tailored perceivers (MLPs) are proposed for each type of signals on the screen, \textit{i.e.}, textual, graphical, and spatial relationship between them, named Textual Perceiver (TxP) $\Phi_{\mathbb{T}}$, Graphical Perceiver (GaP) $\Phi_{\mathbb{G}}$ and Spatial Perceiver (SaP) $\Phi_{\mathbb{S}}$, respectively. We design a specific training recipe to guide each perceiver in extracting modality-specific signals from visual clues $\mathcal{I}$, which will be discussed in \cref{sec: multi_stage_training}.
Thus, we can extract the textual signal $X_t\!=\!\Phi_{\mathbb{T}}(\mathcal{I})$, the graphical signal $X_g \!=\! \Phi_{\mathbb{G}}(\mathcal{I})$, and the spatial signal $X_s\!=\! \Phi_{\mathbb{S}}(\mathcal{I})$ from $\mathcal{I}$, where $X_t,X_g,X_s\in \mathbb{R}^{N \times \mathcal{D}}$. 

\noindent\textbf{Fusion Gate.} 
Different GUI-related downstream tasks exhibit varying preferences for screen content. For example, in widget captioning~\cite{widgetcaption}, the model should pay more attention to graphical signals (icon/widget) related to the question; in screen summarization~\cite{screen2words}, both textual and graphical signals across the screen should be emphasized; and in screen question answering~\cite{screenQA,screenai} and navigation~\cite{aitw,mind2web}, all three signals should be taken into consideration. Thus, the GUI signals extracted by three perceivers should be dynamically fused based on semantics to align with the requirements of tasks. Inspired by MoE~\cite{zong2024movaadaptingmixturevision, lin2024moe, dai2024deepseekmoe} technology, we introduce FG to achieve the above goals.

Specifically, we first concatenate three perceiver signals $X_t$, $X_g$ and $X_s$ as a union of modality information, $X_f = [X_t;X_g;X_s]$, 
where $X_f\in \mathbb{R}^{3N \times \mathcal{D}}$ and $';'$ means concatenation operation. Subsequent to a self-attention operation, interacting $\mathcal{Q}$ with $\overline{\mathcal{I}}$ to align dimensions and enhance the semantics of $\mathcal{Q}$ results in the gating signal $\mathcal{G}\in \mathbb{R}^{N \times \mathcal{D}}$. Next, $\mathcal{G}$ is used to fuse features with $X_f$, incorporating semantic awareness. We can formulate this as follows:
\begin{equation} \label{formula:attention1}
\mathcal{G} = \operatorname{softmax}(\frac{(\overline{\mathcal{I}} W^Q_{g})(\mathcal{Q} W^{K}_{g})^T}{\sqrt{\mathcal{D}}})(\mathcal{Q} W^V_{g}),
\end{equation}
\begin{equation} \label{formula:attention2}
\mathcal{X}_t = \operatorname{softmax}(\frac{(\mathcal{G} W^Q_t)(X_f W^K_t)^T}{\sqrt{\mathcal{D}}})(X_f W^V_t) ,
\end{equation}
 
\noindent where $\mathcal{X}_t \in \mathbb{R}^{N \times \mathcal{D}}$ denotes task-oriented
features, which could serve as GUI-tailored visual clues, and $W^Q_g, W^K_g,W^V_g,W^Q_t,W^K_t,W^V_t\in \mathbb{R}^{\mathcal{D} \times \mathcal{D}}$ are learnable parameter matrices.

With FG, \model interprets the task semantics of the question, interacts with the extracted GUI signals of three perceivers, and dynamically fuses them to produce task-oriented features. Finally, $\overline{\mathcal{I}}$, $\mathcal{X}_t$ and $\mathcal{Q}$ are concatenated along the sequence dimension as LLM input.

 \subsection{Multi-stage Training Strategy} \label{sec: multi_stage_training}

Our TGS-PFM consists of two components: three perceivers (TxP, GaP and SaP) and FG. The perceivers focus on modality-specific signals, while FG dynamically fuses these signals based on task semantics to produce GUI-tailored visual clues. The order of training these components is crucial, hence we introduce a Multi-stage Training Strategy (MTS).
\cref{table:msts} presents the overview of MTS.

\noindent \textbf{Perceivers Training.} We train each perceiver (\textit{Step 1–3}) using specialized data to guide TGS-PFM to extract modality-specific signals on the screen. We first train TxP and GaP, aiming to warm up \model in adapting GUI-related scenarios. Afterwards, we train the SaP, which implicitly relies on TxP and GaP to model spatial relationships between screen elements. 

\noindent \textbf{FG Training.}
After training perceivers, each of them has the ability to be aware of modality-specific signals (textual, graphical, and spatial content) on the screen. To equip FG with the ability to interpret task preferences from the question semantics and dynamically fuse each type of signal, we mix the training data with multiple task preferences in the \textit{Step 4} training stage.

After training, the TGS-PFM is able to perceive and extract distinct GUI signals from the visual clues, and dynamically complete feature fusion based on task semantics.

\section{Data Preparation}
\label{sec:data}

\begin{table}[t]\LARGE
    \centering
    \renewcommand{\arraystretch}{1} 
    \resizebox{0.93\linewidth}{!}{
    \begin{tabular}{ccccc}
        \toprule
        \makecell{\textbf{Training}} & \textbf{Dataset} & \textbf{Task} & \textbf{Samples} & \makecell{\textbf{Unlocked Params}}  \\ \midrule
        \textit{Step 1} & TAD & \makecell{text2bbox\\bbox2text} & 160K & \makecell{TxP\\LLM}   \\
        \midrule
        \textit{Step 2} & GAD & \makecell{text2bbox\\bbox2text} & 187K & \makecell{GaP \\LLM}    \\
        \midrule
        \textit{Step 3} & SAD & SRP     & 200K & \makecell{SaP \\LLM}   \\
        \midrule
        \multirow{5}{*}{\textit{Step 4}}   & TAD & text2bbox   & \multirow{3}{*}{\textit{35K}}  & \multirow{5}{*}{\makecell{TGS-PFM\\LLM\\vision backbone} } \\
                                            & GAD & bbox2text   &    &  \\
                                            & SAD & SRP   &    &  \\
        \cline{2-4}
               & \multirow{2}{*}{\textit{SynD}} &  SPE-QA & \multirow{2}{*}{\textit{48K}} &     \\
               &  &  MPE-QA &    &  \\
        \midrule
               &    &             \textbf{Total} & 680K    & \\
        \bottomrule
    \end{tabular}
    }
    \caption{Details of our Multi-stage Training Strategy (MTS). SynD refers to Synthetic Data (\cref{sec:synthetic_data_generation}), while SRP, SPE-QA, and MPE-QA denote Spatial Relationship Prediction (\cref{manual_adjustment}), Single Perceiver Enhanced QA, and Multi-Perceiver Enhanced QA (\cref{sec:synthetic_data_generation}), respectively. During training, we employ LoRA~\cite{lora} to fine-tune the LLM or the vision backbone.}
    \label{table:msts}
\end{table}

\subsection{Post-Processing Based on Existing Data}\label{manual_adjustment}
In TGS-PFM, TxP extracts text signals from the screen, GaP captures graphical signals, and SaP identifies the spatial relationships between screen elements. To ensure the functional distinctiveness of each perceiver, we constructed specific training data for them.

\noindent\textbf{\textit{Text Aware Data} (TAD).} 
As \textit{Step 1} in \cref{table:msts}, we create TAD to guide the TxP to focus on textual signals within screens. We filter grounding task samples from the AMEX~\cite{amex} dataset by using OCR tools to identify bounding boxes that contain only text. These purely textual targets are retained as \textit{text2bbox} pairs\footnote{All boxes related are defined by $[x_{left}, y_{top}, x_{right}, y_{bottom}]$, scaled to $[0, 1000]$.}. Further, we exchange the function description and coordinates as \textit{bbox2text} data.

\noindent\textbf{\textit{Graphics Aware Data} (GAD).} In \textit{Step 2}, to focus GaP on graphical information, \textit{e.g.}, icons and widgets, we apply the same method of TAD, selecting samples without text in bounding box. Additionally, \textbf{we find that MLLMs struggle to perceive fine-grained visual information on screens.} To alleviate this weakness, we created a grounding dataset for small objects derived from AITW~\cite{aitw}. Specifically, we treat episodes as independent QA pairs and calculate the screen proportion of the target element within each pair using the formula $r = \frac{w\times h}{\mathit W \times \mathit H}\!\times \!100\%$, where $w$ and $h$ represent the width and height of the box, and $\mathit W$ and $\mathit H$ represent the resolution of the screen image. Ultimately, we retain only those targets for which $r <= 0.3\%$.


\begin{figure}
  \centering
  \includegraphics[width=1.\linewidth]{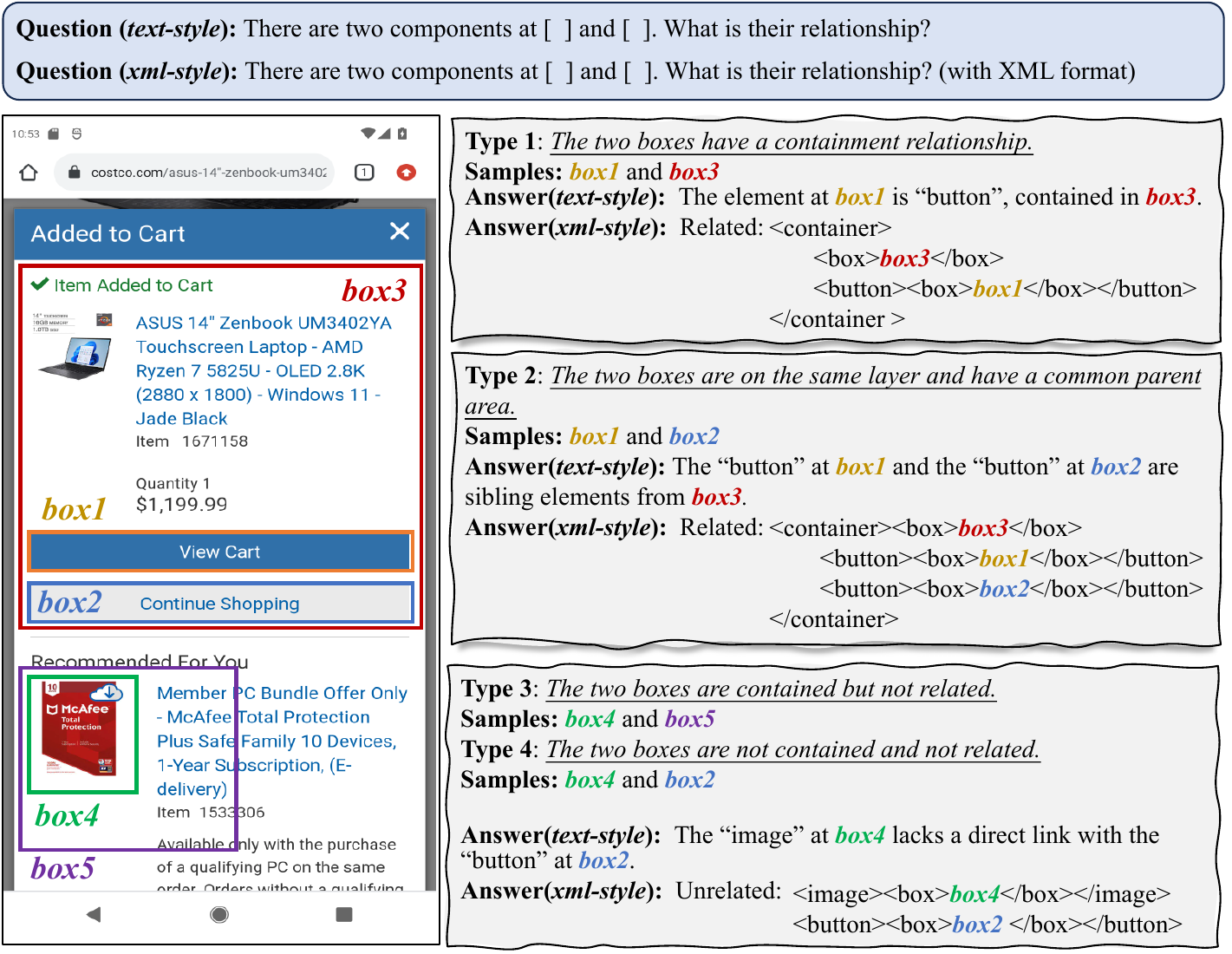}
  \caption{Examples of Spatial Relationship Prediction (SRP) task.}
  \label{fig:SRP}
\end{figure}

\noindent\textbf{\textit{Spatial Aware Data} (SAD). }
Previous research has focused on acquiring structured layout knowledge on the screen via explicit provision of code-like clues to the model~\cite{li2020mapping,li2021vut,pix2struct} or via autoregressive reconstruction of the screen code to implicitly guide model learning~\cite{screenai}. \textbf{We introduce a novel task called Spatial Relationship Prediction (SRP), which explicitly models the spatial relationships between elements on the screen to guide \model in clearly perceiving the spatial context of GUI elements. }

As depicted in \cref{fig:SRP}, the SRP task comprises four types. \textit{Type 1}: containment relationships (one element is the parent node of another in the VH tree). \textit{Type 2}: two elements share the same parent node and depth. 
\textbf{To prevent the model from making predictions based solely on the box coordinates in question}, we introduce two types of negative samples. \textit{Type 3}\footnote{We randomly select the elements and expand their boxes to generate negative samples. We limit the expanded box to have $\operatorname{IoU}\leq0.1$ with the original box and $\operatorname{IoU}\leq0.3$ with the box of its parent node in the VH tree.}: two boxes have a containment relationship but no actual association with UI design. \textit{Type 4}\footnote{These pairs are randomly chosen from the VH tree file except for those included in \textit{Type 1} and \textit{Type 2}.}: two boxes without either containment or association. 

These tasks allow the SaP to perceive the visual, semantic and spatial relationships of screen elements more effectively. The SRP dataset is constructed using the original screen code files of images from the Semantic UI dataset~\cite{rico_semantic}. For \textit{Type 1} and \textit{Type 2}, the root node is excluded, and only 1-hop connections within the VH tree are retained to ensure strong semantic and spatial correlations among boxes. More details of SRP are in Suppl.Mater.

\subsection{Data Synthesis via MLLM}
\label{sec:synthetic_data_generation}
Unlike the vanilla MoE~\cite{lin2024moe,dai2024deepseekmoe,jiang2024mixtralexperts}, which uses a router to control the information flow to experts, we introduce a task-oriented FG to dynamically fuse the output GUI-related signals according to task semantics (\cref{sec:model_architecture}). To support FG training, different tasks with clear semantic preferences should be mixed together; thus, we propose a data synthesis pipeline with an advanced MLLM (Qwen2-VL-72B~\cite{Qwen2VL}). Specifically, we prompt the MLLM to generate two types of synthetic data, namely, \textbf{\textit{Single Perceiver Enhanced Question Answering} (SPE-QA)} and \textbf{\textit{Multi-Perceiver Enhanced Question Answering} (MPE-QA)}. 

For \textbf{SPE-QA}, the data should have a strong semantic preference for one modality of content on the screen, such as OCR-related QA pairs (for TxP) and graphics-related captioning (for GaP) tasks. Since each GUI perceiver is pre-trained on specific type of data (\cref{manual_adjustment}) and are therefore able to coarsely distinguish tasks with different semantics, we are hopeful that \textbf{SPE-QA} data can guide FG to selectively enhance the fusion of specific GUI signals.

The role of \textbf{MPE-QA} data is to further improve FG's semantic understanding and weight assignment capabilities by introducing semantic integration tasks that require FG to have the ability to collaborate with different perceivers. For this type of task, we prompt the MLLM to provide a fine-grained description of the screen content\footnote{We found that Qwen2-VL-72B~\cite{Qwen2VL} effectively performs OCR and has coarse but useful graphics perception. While its quality isn't on par with human supervision, it is still worth considering for training FG.}. The generated output includes textual content, graphical elements, and coarse-grained positional information of elements. In order to synthesize multi-granularity screen perception data, we further introduce a task known as \textbf{\textit{Local Description}}. This task focuses on capturing the contextual information surrounding graphics to enhance local cognition. Specifically, we utilize an enhanced version of YOLO\footnote{We fine-tuned YOLOv8 using about 9k manually labeled in-house data, which mainly includes medical appointment registration, catering, funds, logistics and other scenes in Alipay.
} to detect graphics on the screen, draw their bounding boxes on the image (similar to the Set-of-Marks~\cite{setofmark} method), and subsequently input the image into the MLLM. We then prompt the MLLM to consider the surrounding content of the target area as much as possible when generating descriptions, thereby facilitating the inclusion of contextual associations in the output. See Suppl.Mater for more details.

\section{Experiments} 
\label{sec:experiments}
\begin{table*}[!ht]
\centering
\begin{minipage}[t]{0.73\columnwidth} 
    \centering
    \resizebox{\linewidth}{!}{
    \begin{tabular}{@{}lll@{}}
    \toprule
    \textbf{Benchmark}                 & \textbf{Metric} &  \\ 
    \midrule
    \textbf{Screen Analysis}           &                 &  \\
Widget Captioning(WC)~\cite{widgetcaption}                & CIDEr           &  \\
Taperception(TP)~\cite{tap}                    & F1              &  \\
\textbf{Screen Question-Answering} &                 &  \\
ScreenQA(QA)~\cite{screenQA}                           & ROUGE-L        &  \\
ScreenQA Short(QAS)~\cite{screenai}                     & SQuAD F1        &  \\
Complex ScreenQA(CQA)~\cite{screenai}                   & SQuAD F1        &  \\
WebSRC(WS)~\cite{chen2021websrc}                             & SQuAD F1        &  \\
\textbf{Screen Grounding}         &                 &  \\
RefExp(RE)~\cite{refexp}                             & Acc@IoU=0.1     &  \\
\textbf{Screen Summarization}      &                 &  \\
Screen2Words(S2W)~\cite{screen2words}                       & CIDEr           &  \\ 
\bottomrule
\end{tabular}
}
\caption{Details of basic GUI understanding benchmark and their metric. The abbreviations of these datasets are given in brackets.}
\label{table:benchmark}
\end{minipage}
\hfill
\begin{minipage}[t]{1.27\columnwidth}
\centering
\renewcommand{\arraystretch}{1.1} 
\setlength{\tabcolsep}{0.05cm}
\resizebox{1\linewidth}{!}{
\begin{tabular}{lccccccccc}
\toprule
 \textbf{Method} & \textbf{Size} & \textbf{WC} & \textbf{S2W} & \textbf{RE} & \textbf{TP} & \textbf{WS} & \textbf{QA} & \textbf{QAS} & \textbf{CQA} \\
\midrule
Qwen-VL~\cite{Qwen-VL} & 9.6B &84.1&	100.2&	36.6	&83.5	&57.3&	78.9&	69.1&	54.9  \\
MiniCPM-V 2.6~\cite{yao2024minicpm} & 8B & 110.5 & 107.3 & 48.5 & 80.4 & 85.2 & 76.3 & 77.3 & 71.5 \\
Qwen2-VL~\cite{Qwen2VL} & 7B &136.6&	98.9&	47.6&	88.0&	82.8&	87.0&	87.8	&65.3 \\
Llama 3.2-V~\cite{Llama3} & 11B & 113.6 & 108.8 & 51.3 & 83.4 & 87.0 & \underline{88.4} & \textbf{91.6} & 74.6 \\
CogAgent~\cite{cogagent} & 18B &136.2&	115.0&	\underline{73.3}&	\textbf{88.4}&	63.1&	85.3&	74.6&	65.1  \\
InternVL2~\cite{chen2024far} & 8B & 140.6&	\underline{115.2}&	71.7&	86.7&	\textbf{89.7}&	84.2&	89.2&	\underline{82.4} \\
InternVL2-P & 8B & \underline{142.8}& 115.1& 72.4& 87.8& \underline{89.6}& 87.2& 88.9&81.9\\
\midrule
\model & 8B & \makecell{\textbf{151.0}\\ \small{$+5.7\%$}}&	\makecell{\textbf{118.4}\\ \small{$+2.8\%$}}&	\makecell{\textbf{83.0}\\ \small{$+13.2\%$}}&	\makecell{\underline{88.2}\\ \small{$-0.2\%$}}&	\makecell{89.2\\ \small{$-0.6\%$}}&	\makecell{\textbf{88.6}\\ \small{$+0.2\%$}}&	\makecell{\underline{90.5}\\ \small{$-1.2\%$}}&	\makecell{\textbf{84.3}\\ \small{$+2.3\%$}}\\
\bottomrule
\end{tabular}
}
\captionof{table}{Comparison of \model with other advanced MLLMs. \textbf{Bold} represents the best results, \underline{underlined} represents the second best results, and the improvement of each task is compared with the second best method. We conduct multi-task fine-tuning for all MLLMs individually on the basic GUI understanding benchmark.}
\label{table:comparation_results}
\end{minipage}
\end{table*}
\begin{table}[]\large
\renewcommand{\arraystretch}{1.} 
\setlength{\tabcolsep}{0.08cm}
\resizebox{\linewidth}{!}{
\begin{tabular}{c|cccc|c}
\toprule
\textbf{Method}  & ScreenAI~\cite{screenai}      & Spotlight~\cite{spotlight}  & Ferret-UI~\cite{you2025ferret} & Pix2Struct~\cite{pix2struct} & \model      \\
\midrule
\textit{\textbf{\#Samples} } & 383.5M        & 2.69M      & 0.84M     & 80M        & 0.68M          \\
\midrule
\textbf{WC}      & \underline {156.4}   & 141.8      & 142.0     & 136.7      & \textbf{156.5} \\
\textbf{S2W}     & \underline {120.8}   & 106.7      & 115.6     & 109.4      & \textbf{121.4} \\
\textbf{RE}      & \textbf{86.3} & -          & -         & -          & \underline {84.7}     \\
\textbf{TP}      & -             & \underline {88.4} & 78.4      & -          & \textbf{88.7}  \\
\textbf{WS}      & \underline {87.2}    & -          & -         & -          & \textbf{90.1}  \\
\textbf{QA}      & \textbf{91.9} & -          & -         & -          & \underline {88.7}     \\
\textbf{QAS}     & \textbf{94.6} & -          & -         & -          & \underline {92.7}     \\
\textbf{CQA}     & -    & -          & -         & -          & \textbf{87.7} \\
\bottomrule
\end{tabular}
}
\caption{Comparison of \model with GUI-specific methods. The results we show are from single-task fine-tuning and \textbf{\textit{\#Samples}} means the count of GUI-related training data used in each method.}
\label{table:special_models_comparation}
\end{table}

\subsection{Experimental Setting} \label{sec: experimental_setting}

\noindent \textbf{Basic GUI Understanding Benchmark.}
As described in \cref{table:benchmark}, we extensively collect multiple public GUI-related datasets, including screen summarization~\cite{screen2words}, widget captioning~\cite{widgetcaption}, clickable prediction~\cite{tap}, grounding~\cite{refexp}, question answering~\cite{screenQA,screenai,chen2021websrc}, aiming to comprehensively evaluate the basic GUI understanding ability of MLLMs. Details of the above datasets are in Suppl.Mater.

\noindent\textbf{Screen Navigation Benchmark.}
Compared to basic GUI understanding tasks, screen navigation (\textit{i.e.}, autonomous agent~\cite{seeclick,lin2024showui}) is more challenging as it requires MLLM to have solid GUI understanding, decompose the user's goal into a series of subtasks to be completed as a trajectory, and continuously interact with different screenshots via specific operations like clicking, sliding, and typing. We select AITW~\cite{aitw} (mobile) and Mind2Web~\cite{mind2web} (website) as benchmark, and treat screen navigation as a purely visual problem, using prompt, action spaces and data splitting consistent with previous literature~\cite{seeclick}.

\noindent\textbf{Screen Grounding Benchmark.} To evaluate the grounding ability of MLLMs on devices with different resolutions, in addition to RefExp~\cite{refexp}, we also use ScreenSpot~\cite{seeclick}, which contains grounding tasks for text objects and icon/widget objects in three scenarios: mobile, desktop, and website.

\noindent \textbf{Training Details.}
We initialize the Alignment Projector, vision backbone (InternViT-300M), and Word Embedding Layer\&LLM (InternLM2.5-7B-chat) from InternVL2-8B~\cite{chen2024far}, while training other modules (TGS-PFM) from scratch. Adhering to our MTS procedure (\cref{sec: multi_stage_training}), we set the learning rates at 1e-5 for \textit{Step 1–3}, 5e-6 for \textit{Step 4}, and 5e-4 for benchmark fine-tuning. LoRA~\cite{lora} was applied to both the LLM and vision backbone, with rank and alpha values of 8 and 16 for \textit{Steps 1–3}, and 64 and 128 for \textit{Steps 4} and benchmark fine-tuning. Each stage involved training for 1 epoch using the AdamW optimizer, with baseline MLLMs following their official fine-tuning protocols, also for 1 epoch of multi-task fine-tuning utilizing LoRA~\cite{lora}\footnote{Our multi-task fine-tuning on the basic GUI understanding benchmark reveals that: ($i$) baseline methods achieve optimal performance after 1 epoch, while additional training generally don't show further performance gains; ($ii$) fine-tuning using LoRA outperforms full parameter fine-tuning.}.


For screen navigation, we apply LoRA~\cite{lora} with rank and alpha value set at 128 and 256, a learning rate of 2e-5, and perform 3 epochs on AITW~\cite{aitw} and Mind2Web~\cite{mind2web}. All experiments were performed on 8 A100 GPUs with a global batch size of 64.

\subsection{Main Results}  \label{sec: evaluation_on_benchmark}
\noindent\textbf{Gains of Instruction-tuning for Domain-specific Data.} Instruction-tuning a MLLM using domain-specific or task-related data can enable the model to quickly learn domain knowledge~\cite{seeclick}. We instruction-tune the vanilla InternVL2 \cite{chen2024far} using the GUI-related training data in \cref{table:msts} to obtain InternVL2-P, which can be considered as a GUI knowledge enhancement in-domain model. Next, we keep InternVL2-P with the same fine-tuning setting as \model on each benchmark (\cref{sec: experimental_setting}). 

\begin{figure*}[h]
\begin{minipage}[]{0.86\columnwidth}
  \centering
  \includegraphics[width=1.0\linewidth]{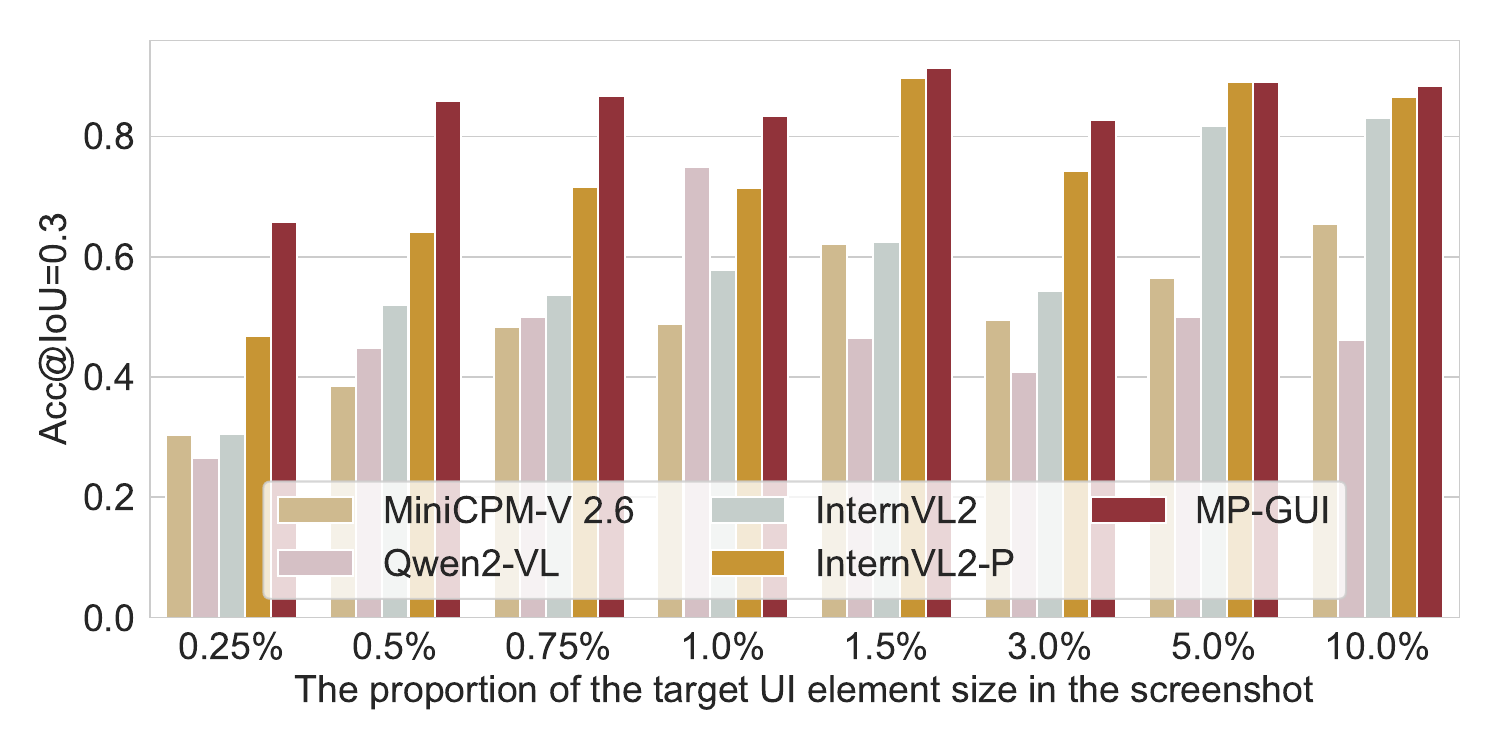}
  \caption{Comparison of the grounding results of various methods on UI elements of different sizes under RefExp~\cite{refexp}. The \textit{proportion} of $k\%$ indicates that $\frac{w\times h}{W \times H} \leq k\%$, where $w$ and $h$ represent the width and height of UI elements, and $W$ and $H$ represent the screen resolution.}
  \label{fig:bar_acc}
\end{minipage}
\hfill
\begin{minipage}[]{1.2\columnwidth}
    \centering
    \renewcommand{\arraystretch}{1.1} 
    \setlength{\tabcolsep}{0.05cm}
    \resizebox{\linewidth}{!}{
    \begin{tabular}{@{}lcccccccc@{}}
        \toprule
        \multirow{2}{*}{\textbf{Method}} & \multirow{2}{*}{\textbf{Size}} &  \multicolumn{2}{c}{\textbf{Mobile}} & \multicolumn{2}{c}{\textbf{Desktop}} & \multicolumn{2}{c}{\textbf{Web}} & \multirow{2}{*}{{Avg.}} \\ 
        \cline{3-4} \cline{5-6} \cline{7-8}
        &  &{Text}&{Icon/Widget}&{Text}&{Icon/Widget}&{Text}&{Icon/Widget}& \\ 
        \midrule
        Llama 3.2-V~\cite{Llama3}&11B&14.7\%&5.7\%&9.3\%&4.3\%&4.3\%&4.4\%&7.1\%\\
        GPT-4V\cite{gpt4v}$^\dag$ & - & 22.6\% & 24.5\% & 20.2\% & 11.8\% & 9.2\% & 8.8\% & 16.2\% \\
        Fuyu~\cite{fuyu}$^\dag$& 8B & 41.0\% & 1.3\% & 33.0\% & 3.6\% & 33.9\% & 4.4\% & 19.5\% \\
        \rowcolor{myrose}
        InternVL2~\cite{chen2024far} & 8B & 74.0 \% & 25.8\% & 54.6\% & 27.1\% & 38.3\% & 31.6\% & 41.9\% \\
        CogAgent~\cite{cogagent}$^\dag$& 18B & 67.0\% & 24.0\% &\textbf{74.2}\% & 20.0\% & \textbf{70.4}\% & 28.6\% & 47.4\% \\
        SeeClick~\cite{seeclick}$^\dag$ & 9.6B & 78.0\% & 52.0\% & 72.2\% & 30.0\%& 55.7\% & 32.5\% & 53.4\% \\
        \hline
        InternVL2-P & 8B & 83.2\%	&52.0\%	&63.4\%	&43.6\%	&47.0\%	&41.3\%	&55.1\% \\
        \rowcolor{myrose}
        \model & 8B & \textbf{86.8\%} & \textbf{65.9\%} & 70.8\% & \textbf{56.4\%} & 58.3\% & \textbf{46.6\%} & \textbf{64.1\%}\\
        \bottomrule
    \end{tabular}
    }
    \captionof{table}{Zero-shot grounding performance on ScreenSpot~\cite{seeclick}. The best results in each column are \textbf{bold}. \dag\ means the results come from SeeClick~\cite{seeclick} and \colorbox{myrose}{pink} color indicates the method we compare.}
    \label{tab:zero_shot}
\end{minipage}
\end{figure*}

It can be observed that for both screen grounding (\cref{fig:bar_acc} and \cref{tab:zero_shot}) and screen navigation (\cref{table:AITW-benchmark} and \cref{table:Mind2Web-benchmark}), InternVL2-P has achieved considerable gains compared to InternVL2~\cite{chen2024far}. However, in \cref{table:comparation_results}, for tasks like CQA and QAS that rely on the foundational model's reasoning abilities, InternVL2-P is inferior to InternVL2~\cite{chen2024far}. Indicating that instruction-tuning may put MLLMs at risk of weakened generic abilities. Notably, \model shows favorable performance over InternVL2-P in all the above benchmarks. 

According to the above results, we argue that \textbf{relying solely on domain-specific data for instruction-tuning enables the MLLM to perceive only superficial GUI information}. Our special designs, which provide additional GUI-tailored visual clues, effectively enhance MLLM in solving GUI-related tasks.

\noindent\textbf{Basic GUI Understanding.} In \cref{table:comparation_results}, \model outperforms baselines on most tasks and ranks second on TP, WS, and QAS. Notably, our \model exceeds the second-best methods by $5.7\%$ on WC and $13.2\%$ on RE, which demonstrates the excellent spatial location awareness of elements on the screen. Compared to Llama 3.2-V(11B)~\cite{Llama3} and CogAgent(18B)~\cite{cogagent}, \model remains competitive with less parameters, exhibiting only a decrease of $0.2\%$ and $1.2\%$ on TP and QAS tasks. In \cref{table:special_models_comparation}, we list current advanced GUI-specific methods~\cite{screenai,spotlight,you2025ferret,pix2struct}. 
Compared with these methods that require a large amount of GUI-related pre-training data (\textit{e.g.}, close-sourced ScreenAI~\cite{screenai} with 383.5M pre-training samples), our method leveraging only 0.68M GUI-specific training samples outperforms them on most tasks and remains competitive in others.

These results show that \model has advanced GUI understanding capabilities, especially in tasks with graphical content (\textit{e.g.}, WC, RE, and CQA). Our special model architecture and training recipe (\textit{i.e.,} MTS) can guide MLLM to learn GUI knowledge effectively with limited data.

\noindent \textbf{Screen Grounding.} The significance of MLLMs' grounding for GUI tasks was explored in prior literature~\cite{seeclick}. Here, we comprehensively evaluate the grounding ability of \model from: \textbf{\textit{(i)}} \textbf{fine-grained perception of small-sized objects}; \textbf{\textit{(ii)}} \textbf{perception of text and graphical objects (icon/widget) on different-resolution devices}.

In \cref{fig:bar_acc}, it is evident that most MLLMs struggle to accurately ground UI elements of smaller size. Thanks to the training on GAD data (which includes \textit{text2bbox} and \textit{bbox2text} tasks involving custom small-size icons/widgets) and our dual-visual-clues framework (\cref{manual_adjustment}), \model achieves the best performance especially for the small-sized objects (with a \textit{proportion} $\leq 1\%$). Notably, InternVL2-P has made significant gains compared to InternVL2~\cite{chen2024far}, verifying the effectiveness of the domain-specific data we constructed. However, \textbf{on small-size objects, InternVL2-P is still weaker than \model, demonstrating the effectiveness of our model aspect optimization.}

In \cref{tab:zero_shot}, our \model consistently attains competitive results in zero-shot grounding across different device scenarios~\cite{seeclick}. Notably, it shows leading performance in grounding graphical objects (icon/widget), achieving significant gains compared to InternVL2~\cite{chen2024far}. Moreover, our training data only contains mobile resolution images (\cref{sec:data}), and do not use website and desktop resolution images to provide priors~\cite{seeclick,lin2024showui}, showing that \textbf{there are generic GUI patterns across different device scenarios our \model can perceive}.

\begin{table}[ht]
\renewcommand{\arraystretch}{1.} 
\resizebox{1.0\linewidth}{!}{
\begin{tabular}{lcccccc}
\toprule
\textbf{Method}  & {General} & {Install} & {G.Apps} & {Single} & {WebShop} & {Overall}      \\
\midrule
GPT-4V~\cite{yan2023gpt} & 41.7 & 42.6 & 49.8 & 72.8 & 45.7 & 50.5 \\
Qwen-VL~\cite{Qwen-VL} & 49.5 & 59.9 & 46.9 & 64.7 & 50.7 & 54.3 \\
OmniParser~\cite{lu2024omniparser} & 48.3 & 57.8 &51.6 & \textbf{77.4} & 52.9 & 57.7 \\
SeeClick~\cite{seeclick} & 54.0 & 66.4 & 54.9 & 63.5 & 57.6 & 59.3 \\
\rowcolor{myrose}
InternVL2~\cite{chen2024far} & 58.1 & 65.3 & 56.8 & 68.7 & 61.1 & 62.0\\
ShowUI~\cite{lin2024showui} & \underline{63.5} & \underline{72.3} & \textbf{66.0} & 72.3 & \underline{65.8} & \underline{68.3} \\
\midrule
InternVL2-P & 61.2&	70.3&	61.6&	74.6	&65.1	&66.6 \\
\rowcolor{myrose}
\model & \textbf{63.7} & \textbf{74.3} & \underline{65.3} & \underline{75.4} & \textbf{67.2} & \textbf{69.2} \\
\bottomrule
\end{tabular}
}
\caption{Performance of Screen Navigation on AITW~\cite{aitw}. The \colorbox{myrose}{pink} color indicates the method we compare.}
\label{table:AITW-benchmark}
\end{table}

\begin{table}[ht]
    \centering
    \renewcommand{\arraystretch}{1.} 
    \setlength{\tabcolsep}{0.07cm}
    \resizebox{1.0\linewidth}{!}{
    \begin{tabular}{@{}lccccccccc@{}}
        \toprule
        \multirow{2}{*}{\textbf{Method}}  &  \multicolumn{3}{c}{{Cross-Task}} & \multicolumn{3}{c}{Cross-Website} & \multicolumn{3}{c}{{Cross-Domain}} \\ 
        \cline{2-4} \cline{5-7} \cline{8-10}
        &  {Ele.Acc}&{Op.F1}&{Step.SR}&{Ele.Acc}&{Op.F1}&{Step.SR}& {Ele.Acc}&{Op.F1}&{Step.SR}\\ 
        \midrule
        \rowcolor{myrose}
        InternVL2~\cite{chen2024far} & 18.8	& 87.4	&16.7& 17.6& 85.8& 14.5 &13.9&	87.0	&12.0  \\
        CogAgent~\cite{cogagent} & 22.4 & 53.0 & 17.6 & 18.4 & 42.4 & 13.4 & 20.6 & 42.0 & 15.5 \\
        SeeClick~\cite{seeclick} & 28.3 & 87.0 & 25.5 & 21.4 & 80.6 & 16.4 & 23.2 & 84.8 & 20.8 \\
        GPT-4~\cite{gpt4} & \underline{41.6} & 60.6 &36.2 &35.8 & 51.1 & 30.1 & 37.1 & 46.5 & 26.4 \\
        ShowUI~\cite{lin2024showui} & 39.7 & \underline{88.0} & \underline{36.9} & \textbf{41.0} & 83.6 & \textbf{34.2} & \textbf{38.9} & 85.3 & \textbf{34.1} \\
        \midrule
        InternVL2-P & 27.4&	87.8&	24.0&	27.4&	\underline{86.1}&	23.1& 24.3&	\underline{87.1}&	21.1 \\
        \rowcolor{myrose}
        \model & \textbf{42.1} & \textbf{89.0} & \textbf{38.1} & \underline{39.4} & \textbf{87.1} & \underline{32.9} & \underline{37.6} & \textbf{87.4} & \underline{33.7} \\
        \bottomrule
    \end{tabular}
    }
    \caption{Performance of Screen Navigation on Mind2Web~\cite{mind2web}. The \colorbox{myrose}{pink} color indicates the method we compare.}
    \label{table:Mind2Web-benchmark}
\end{table}
\begin{table*}[ht]
    \centering
    \renewcommand{\arraystretch}{1.} 
    \resizebox{0.93\linewidth}{!}{
    \begin{tabular}{lcccccccc}
        \toprule
        \textbf{Method} & \textbf{WC} & \textbf{S2W} & \textbf{RE} & \textbf{TP} & \textbf{WS} & \textbf{QA} & \textbf{QAS} & \textbf{CQA} \\
        \hline
        \textit{w/o} FG$^\dagger$ & 142.1 \textcolor{mygreen}{(\small{-6.3\%})}	&117.8 \textcolor{mygreen}{(\small{-0.5\%})}&	76.8 \textcolor{mygreen}{(\small{-8.1\%})}	&87.9 \textcolor{mygreen}{(\small{-0.3\%})}&	87.2 \textcolor{mygreen}{(\small{-2.3\%})}&	87.3 \textcolor{mygreen}{(\small{-1.5\%})}&	89.1 \textcolor{mygreen}{(\small{-1.6\%})}&	 80.7 \textcolor{mygreen}{(\small{-4.5\%})} \\

        \textit{w/o} FG$^\ddagger$ & 143.4 \textcolor{mygreen}{(\small{-5.3\%})}	&116.7 \textcolor{mygreen}{(\small{-1.5\%})}&	77.5 \textcolor{mygreen}{(\small{-7.1\%})}	&88.1 \textcolor{mygreen}{(\small{-0.1\%})}&	89.3 \textcolor{red}{(\small{+0.1\%})}&	88.0 \textcolor{mygreen}{(\small{-0.7\%})}&	89.3 \textcolor{mygreen}{(\small{-1.3\%})}&	82.6 \textcolor{mygreen}{(\small{-2.1\%})} \\

        \textit{w/o} TxP & 142.6 \textcolor{mygreen}{(\small{-5.9\%})}&	115.2 \textcolor{mygreen}{(\small{-2.8\%})}&	78.8 \textcolor{mygreen}{(\small{-5.3\%})}&	88.3 \textcolor{red}{(\small{+0.1\%})}&	89.1 \textcolor{mygreen}{(\small{-0.1\%})}&	87.5 \textcolor{mygreen}{(\small{-1.3\%})}&	89.4 \textcolor{mygreen}{(\small{-1.2\%})}&	80.8 \textcolor{mygreen}{(\small{-4.3\%})}  \\
 
        \textit{w/o} GaP &  143.1 \textcolor{mygreen}{(\small{-5.5\%})}&	116.0 \textcolor{mygreen}{(\small{-2.1\%})}&	79.5 \textcolor{mygreen}{(\small{-4.4\%})}&	88.1 \textcolor{mygreen}{(\small{-0.1\%})}&	89.3 \textcolor{red}{(\small{+0.1\%})}&	87.7 \textcolor{mygreen}{(\small{-1.0\%})}	&89.5 \textcolor{mygreen}{(\small{-1.1\%})}&	80.3 \textcolor{mygreen}{(\small{-5.0\%})}  \\

        \textit{w/o} SaP & 141.9 \textcolor{mygreen}{(\small{-6.4\%})}	&116.2 \textcolor{mygreen}{(\small{-1.9\%})}	&78.4 \textcolor{mygreen}{(\small{-5.9\%})}	&88.2 (\small{0.0\%})&	89.0 \textcolor{mygreen}{(\small{-0.2\%})}&	87.6 \textcolor{mygreen}{(\small{-1.1\%})}&	89.4 \textcolor{mygreen}{(\small{-1.2\%})}&	80.3 \textcolor{mygreen}{(\small{-5.0\%})}  \\
        \textit{w/o} MTS & 148.3 \textcolor{mygreen}{(\small{-1.8\%})}&117.0 \textcolor{mygreen}{(\small{-1.2\%})}& 82.4 \textcolor{mygreen}{(\small{-0.7\%})}&87.2 \textcolor{mygreen}{(\small{-1.1\%})}& 86.9 \textcolor{mygreen}{(\small{-2.6\%})}&87.4 \textcolor{mygreen}{(\small{-1.4\%})}&88.3 \textcolor{mygreen}{(\small{-2.4\%})}&83.5 \textcolor{mygreen}{(\small{-1.0\%})}\\
        \hline
        \textbf{\model} &  151.0&	118.4&	83.0&	88.2&	89.2&	88.6&	90.5&	84.3\\
        \bottomrule
    \end{tabular}
    }
    \caption{Ablation study results. $\textit{w/o}$ FG$^\dagger$ indicates that we don't pre-train FG (without implementing \textit{Step 4} in the MTS). $\textit{w/o}$ FG$^\ddagger$ indicates that we remove FG and directly use the mean of outputs from the three perceivers for feature fusion. \textit{w/o} MTS means keeping the same settings of \model except that we don't use step-by-step MTS (\cref{sec: multi_stage_training}) but collect all training data for end-to-end training TGS-PFM.}
    \label{tab:ablation}
\end{table*}
\noindent \textbf{Screen Navigation.} The generic abilities of foundational models, such as visual perception and reasoning, are crucial for navigation tasks. In the AITW~\cite{aitw} benchmark (\cref{table:AITW-benchmark}), compared to SeeClick~\cite{seeclick}, which is based on Qwen-VL~\cite{Qwen-VL} and achieved an $8.3\%$ gain, and ShowUI~\cite{lin2024showui}, which is based on Qwen2-VL~\cite{Qwen2VL} and provides a $1.2\%$ improvement, our \model achieves a $10.4\%$ gain compared to the foundational model InternVL2~\cite{chen2024far}. For ShowUI~\cite{lin2024showui}, we choose the version without visual history. In the Mind2Web~\cite{mind2web} benchmark (\cref{table:Mind2Web-benchmark}), \model achieves leading results on action prediction (Op.F1) and competitive results on both click-element location (Ele.Acc) and step success rate (Step.SR). Compared with InternVL2~\cite{chen2024far}, our \model still achieved significant gains.





\subsection{Ablation Study}  \label{sec: ablation}
In \cref{tab:ablation}, we perform ablation studies evaluating (1) the impact of the FG, (2) the effects of different perceivers and (3) the gains of MTS.


\noindent\textbf{Fusion Gate.}
In \cref{sec:model_architecture}, we introduce FG module that dynamically assign weights to different perceivers according to task semantics and global visual signal. In this study, we examine the effect of FG on performance by modifying both the training recipe and the architecture. For \textit{w/o} FG$^\dagger$, the results show that it is necessary to utilize task-oriented data to achieve task-specific semantic awareness of FG. Meanwhile, for \textit{w/o} FG$^\ddagger$, the removal of FG weakens the performance of \model, especially in WC, RE, and CQA tasks, confirming FG's effectiveness.

\noindent\textbf{GUI-tailored Perceivers.} In \cref{tab:ablation}, for S2W, QA, and QAS that prefer to focus on the text information from screens, the gains decrease by $2.8\%/1.3\%/1.2\%$ respectively without TxP. For WC and RE tasks, \model needs to clarify the spatial context among elements. The gains on WC and RE decrease by $6.4\%$ and $5.9\%$ respectively when without SaP. For challenging CQA tasks, it is necessary to comprehensively localize the target element and perceive the spatial contextual information around it to assist MLLM in reasoning. Thus, both GaP and SaP are crucial, and removing either of them will reduce performance by $5.0\%$. 

\noindent\textbf{MTS.} 
We conduct pre-training on TGS-PFM (\cref{sec:model_architecture}) by using the same data (\cref{table:msts}) as that of \model. Instead of adopting our step-by-step MTS recipe (\cref{sec: multi_stage_training}), we opt for the end-to-end training mode. As shown in \cref{tab:ablation}, this end-to-end method results in performance degradation when compared to MTS, validating the gains of MTS.

\begin{figure}
  \centering
  \includegraphics[width=1\linewidth]{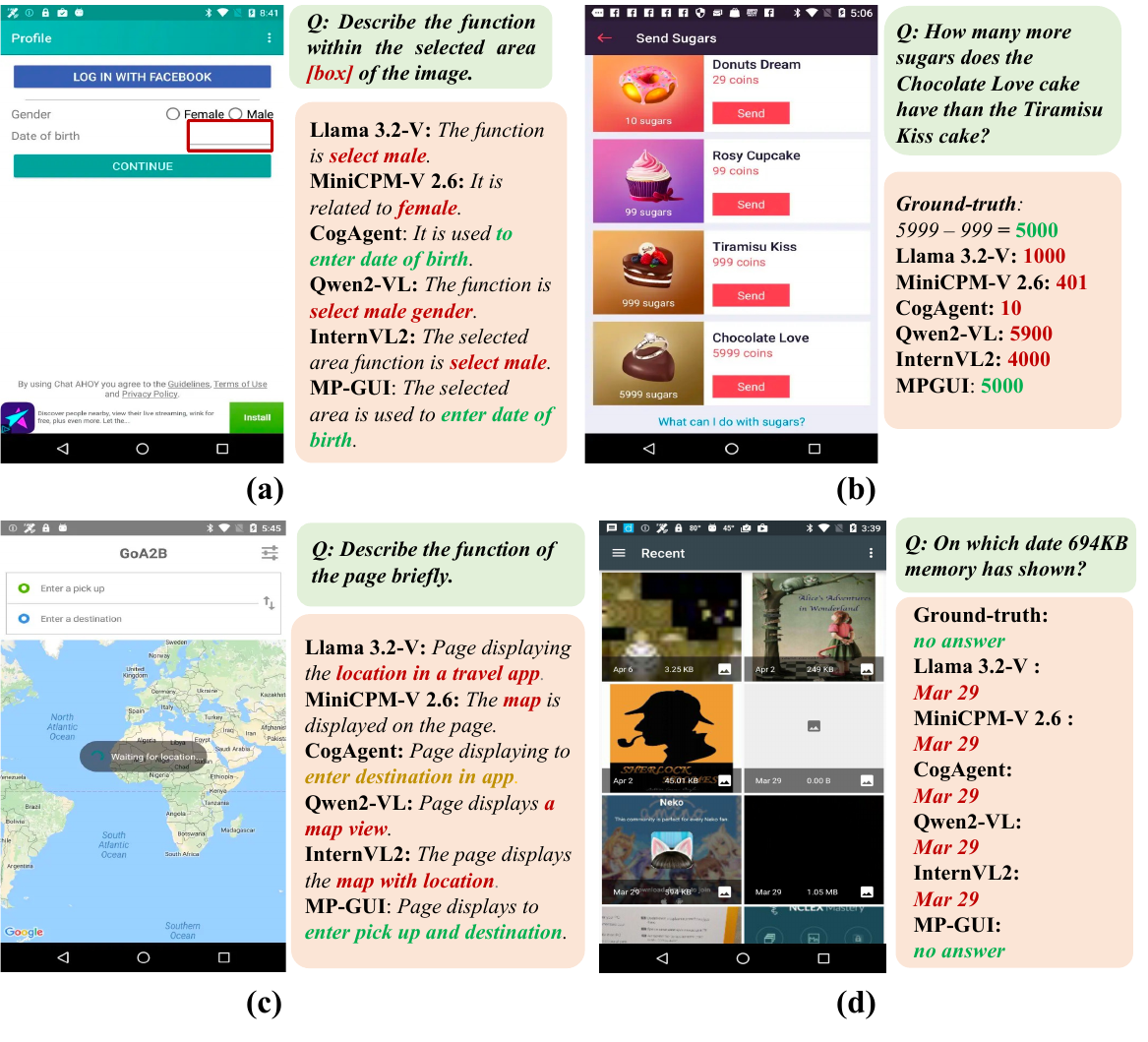}
  \caption{Case studies on basic GUI understanding benchmark (\cref{sec: experimental_setting}). Accurately described answer is marked in \textcolor[rgb]{0.36,0.68,0.21}{\textbf{green}}, while inaccurately and incompletely described ones in \textcolor[rgb]{0.70,0.17,0.13}{\textbf{red}} and \textcolor{orange}{\textbf{orange}}.}
  \label{fig:visual}
\end{figure}

\subsection{Qualitative Analysis}  \label{sec: visualization}
In \cref{fig:visual}, we display several examples on basic GUI understanding tasks (\cref{sec: experimental_setting}). The results demonstrate that our \model effectively understands the implicit knowledge on the screen, including graphics, text, and their spatial relationships (case \textbf{a}). For queries that necessitate deeper reasoning (case \textbf{b}), \model can effectively locates and associates details from different areas to generate accurate answers. Furthermore, \model demonstrates a comprehensive ability to capture essential modality details of the screen (case \textbf{c}). Notably, \model's advanced GUI perception and understanding capabilities can alleviate the hallucination problem often encountered in MLLMs (case \textbf{d}). More results and detailed discussion are in Suppl.Mater.

\section{Conclusion} \label{sec:conclusion}
In this paper, we center on enhancing MLLM in GUI scenarios and present \model, a dual-visual-clues model. It uses three GUI-specific perceivers to extract modality signals from the screen and a FG module to fuse them based on task semantics, generating additional GUI-tailored visual clues to enhance MLLM's GUI visual perception. We also introduce a novel SRP task for explicitly modeling GUI elements' spatial relationships and an automated synthetic data pipeline for FG training. Extensive experiments confirm our designs significantly enhance MLLM's GUI understanding, facilitating the improvement of various downstream tasks.



\textbf{Acknowledgments}
This work was supported by the National Natural Science Foundation of China (Grant No.6237 2408). This work was also supported by Ant Group.
\clearpage
{
    \small
    \bibliographystyle{ieeenat_fullname}
    \bibliography{main}
}
\appendix
\clearpage
\maketitlesupplementary

%
%

Our codes and datasets are publicly available at \textcolor{red}{\textit{\nolinkurl{https://github.com/BigTaige/MP-GUI}}}.

\begin{figure}
  \centering
  \includegraphics[width=\linewidth]{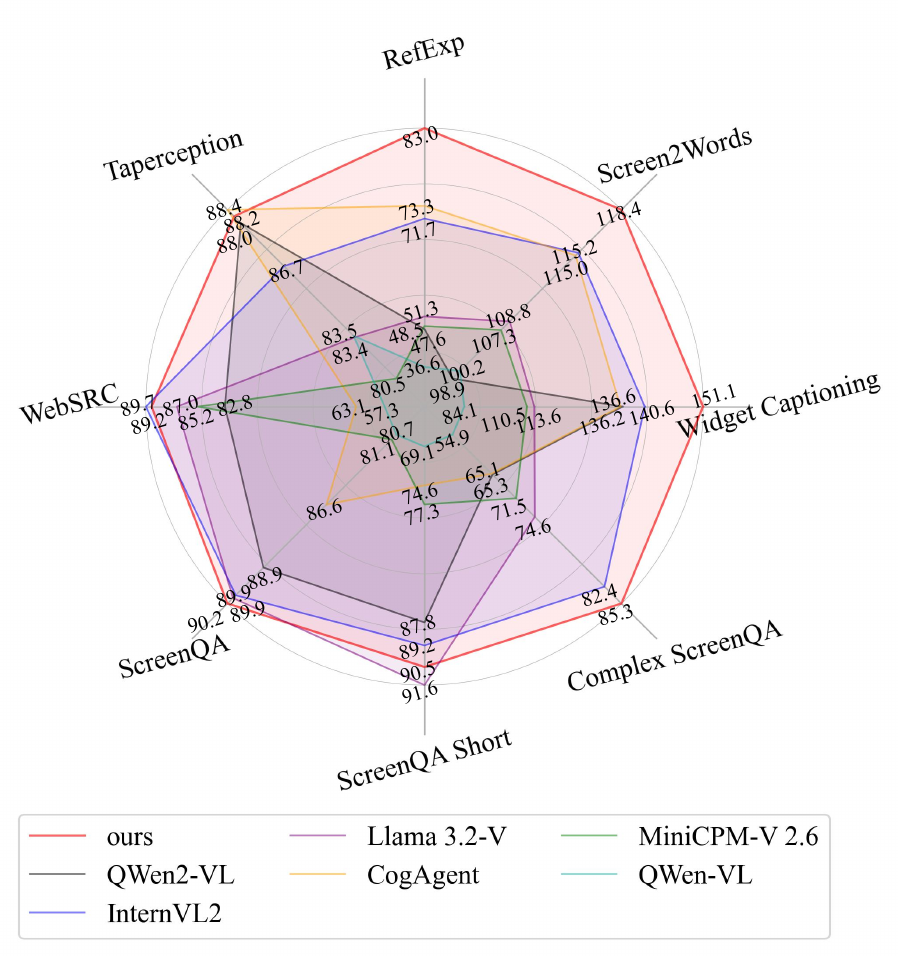}
  \caption{\model outperforms six open-source MLLMs in the GUI understanding benchmark.}
  \label{fig:rader}
\end{figure}

\section{Training Configurations}

We report the detailed settings of \model during multi-step training and multi-task fine-tuning, as shown in \cref{multi_step_config}. 
As introduced in Sec~\ref{sec: multi_stage_training}: \textit{Step 1} represents Textual Perceiver training, \textit{Step 2} represents Graphical Perceiver training, \textit{Step 3} represents Spatial Perceiver training, and \textit{Step 4} is Fusion Gate training.

\begin{table}[h]\LARGE
\centering
\renewcommand{\arraystretch}{1.2}
\resizebox{\linewidth}{!}{
    \begin{tabular}{ccccc|c}
     \toprule
    \textbf{Configurations} & \textit{\textbf{Step 1}} & \textit{\textbf{Step 2}} & \textit{\textbf{Step 3}} & \textit{\textbf{Step 4}} & \textbf{\textit{MFT}} \\
 \toprule
    Training epochs & \multicolumn{5}{c}{1}\\
    
    Max dynamic patch & \multicolumn{4}{c|}{6}& 4\\

    Training samples & 160,031 & 187,657 & 200,000 & 93,419 &107,373\\

    Warmup ratio & \multicolumn{5}{c}{0.03} \\
    
    Warmup decay & \multicolumn{5}{c}{0.01} \\

    Global batch size & \multicolumn{5}{c}{64} \\
    
    Learning rate & \multicolumn{4}{c|}{$1 \times 10^{-5}$}& $4 \times 10^{-5}$  \\

    Learning rate decay & \multicolumn{5}{c}{Cosine schedule}  \\
    Optimizer & \multicolumn{5}{c}{AdamW}  \\
    Adam $\epsilon$ & \multicolumn{5}{c}{$1 \times 10^{-8}$}  \\
    Adam $\beta$ & \multicolumn{5}{c}{$(0.9,0.999)$}  \\
    \bottomrule
    \end{tabular}
}
\caption{Training configuration details. \textbf{\textit{MFT}} means multi-task fine-tuning.}
\label{multi_step_config}
\end{table}

    
    

    

    


\section{Details of Evaluation Datasets}\label{supp_details_benchamrk}
In this section, we describe the details of each task in the GUI understanding benchmark and the templates we used.

\textbf{Widget Captioning (WC) ~\cite{widgetcaption}:} It is a benchmark for automatically generating language description for the functionality of an object on the screen. The numbers of samples for the partitioned train/val/test are 14,878/1,292/1,265 respectively. The template we use is as follows, where \textbf{\textit{bbox}} represents the coordinates area of the target and the \textit{$<$image$>$} is a placeholder that will be replaced by image tokens:

\begin{tcolorbox}[title = {The template for \textbf{Widget Captioning}}]
\textit{$<$image$>$$\backslash$n Describe the function within the selected area $<$box$>$ [\textbf{bbox}] $<$/box$>$ of the image. answer with phrases rather than sentence.}
\end{tcolorbox}

\textbf{Taperception (TP)~\cite{tap}:} This benchmark is used to predict whether a given target element is clickable. It can be used to detect the accessibility of GUI elements on the screen. The numbers of samples for the partitioned train/val/test are 14,781/1,857/2,029. The template employed for this task is as follows:

\begin{tcolorbox}[title = {The template for \textbf{Taperception}}]
\textit{$<$image$>$$\backslash$n Whether the graphic within the selected area $<$box$>$ [\textbf{bbox}] $<$/box$>$ is clickable? If clickable, output 0. otherwise output 1.}
\end{tcolorbox}

\textbf{ScreenQA (QA) ~\cite{screenQA}:} This is a benchmark for screen comprehension. It comprises UI elements and full-sentence answers as the ground truth. The objective of this dataset is to extract the OCR content from the screen in conjunction with the given question. The numbers of samples for the partitioned train/val/test are 68,951/8,614/8,419. The template used is as follows, where \textbf {\textit {question}} represents the original question of the sample.

\begin{tcolorbox}[title = {The template for \textbf{ScreenQA}}]
\textit{$<$image$>$$\backslash$n \textbf{question}?}
\end{tcolorbox}

\textbf{ScreenQA Short (QAS)~\cite{screenai}:} It is a modified version of ScreenQA~\cite{screenQA}, having the same questions for the same screenshots, with answers autogenerated by PaLM 2-S~\cite{chen2023pali} from original human-annotated data. The numbers of samples for the partitioned train/val/test are 68,951/8,614/8,419. The template acting on it is as follows: 

\begin{tcolorbox}[title = {The template for \textbf{ScreenQA Short}}]
\textit{$<$image$>$$\backslash$n \textbf{question}? Answer with numbers or phrases rather than sentence.}
\end{tcolorbox}

\textbf{Complex ScreenQA (CQA)~\cite{screenai}:} An extension or substitute of ScreenQA Short~\cite{screenai}, which incorporates more arduous questions, namely those related to counting, arithmetic, comparison, and non-answerable varieties, as well as screens possessing diverse aspect ratios, is employed to assess the model's proficiency in localizing, spatial perception and reasoning about screen elements, which needs multipart screen information. As the original data lacks details on data division, yet the author noted in the data card that CQA is founded on data synthesized by QAS~\cite{screenai}, in this study, we partition the CQA data in line with the image index in QAS~\cite{screenai}. Finally, the numbers of samples for the partitioned train/val/test are 6,347/796/759. We maintain the template adopted in CQA consistent with that of QAS:

\begin{tcolorbox}[title = {The template for \textbf{Complex ScreenQA}}]
\textit{$<$image$>$$\backslash$n \textbf{question}? Answer with numbers or phrases rather than sentence.}
\end{tcolorbox}

\textbf{WebSRC (WS)~\cite{chen2021websrc}:}
This is a web scenario question-answering benchmark, with the answers primarily centered around the OCR content within the page. The numbers of samples for the partitioned train/val/test are 307,315/4,558/4,558. We ensure that the template remains in line with that of QAS~\cite{screenai}:

\begin{tcolorbox}[title = {The template for \textbf{WebSRC}}]
\textit{$<$image$>$$\backslash$n \textbf{question}? Answer with numbers or phrases rather than sentence.}
\end{tcolorbox}

\textbf{RefExp (RE)~\cite{refexp}:}
This is a task of generating the coordinates of the object referred to in the query, used to evaluate the model's accuracy in locating and identifying the position of specific objects within a given context. The numbers of samples for the partitioned train/val/test are 15,624/471/565. The template utilized on it is as follows, where \textbf{\textit{reference}} represents the description of the target element:

\begin{tcolorbox}[title = {The template for \textbf{RefExp}}]
\textit{$<$image$>$$\backslash$n Please provide the bounding box coordinate of the region this sentence describes: $<$ref$>$ \textbf{reference}$<$/ref$>$}
\end{tcolorbox}

\textbf{Screen2Words (S2W)~\cite{screen2words}:}
This benchmark requires the model to be aware of the global and local information of the screen and use a concise text to summarize the content and function of the current screen. The numbers of samples for the partitioned train/val/test are 15,743/2,364/4,310. The template that we employed for this particular task is as follows:

\begin{tcolorbox}[title = {The template for \textbf{Screen2Words}}]
\textit{$<$image$>$$\backslash$n Use a phrase to describe the function of the page.}
\end{tcolorbox}

For all the above tasks, we format them into conversational QA pairs to adapt to the inference and training mode of MLLMs. To balance data distribution in multi-task fine-tuning, we sample only the first 10,000 samples from the QA~\cite{screenQA} and QAS~\cite{screenai} datasets, and the first 20,000 samples from the WS~\cite{chen2021websrc} dataset.

\section{Analysis of GUI Perceivers}
\begin{figure}[ht]
    \centering
    \begin{subfigure}{\linewidth}
        \includegraphics[width=\textwidth]{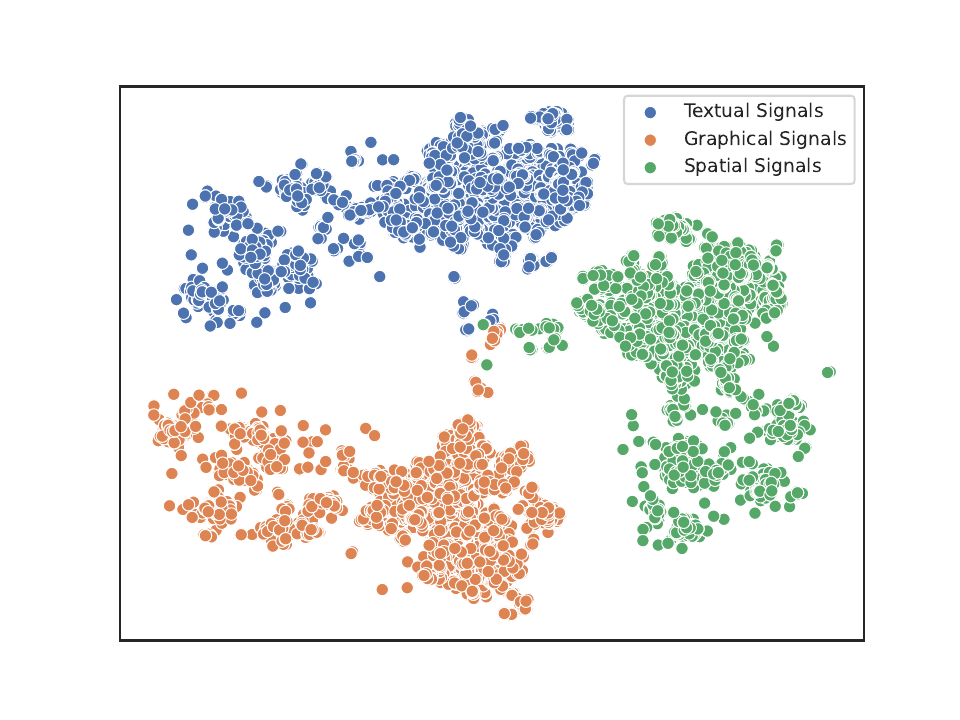}
        \caption{Results on Screen2Words~\cite{screen2words}.}
    \end{subfigure}
    \begin{subfigure}{\linewidth}
        \includegraphics[width=\textwidth]{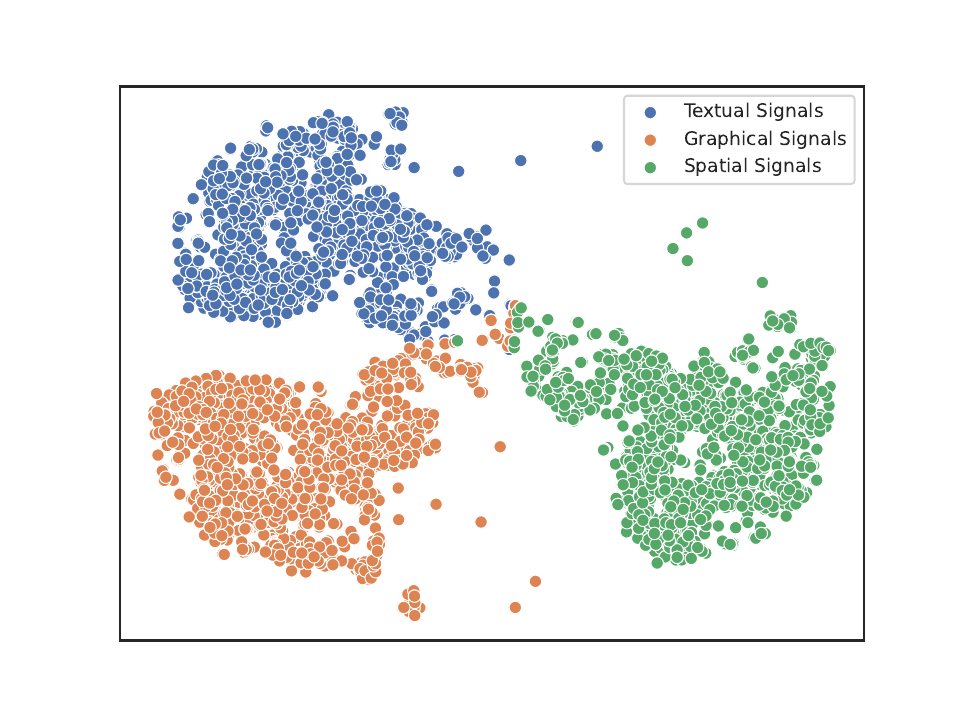}
        \caption{Results on WebSRC~\cite{chen2021websrc}.}
    \end{subfigure}
    \caption{Visualization results of different GUI modality signals processed by t-SNE.}
    \label{visulation_tSNE}
\end{figure}

In this section, to confirm that different GUI Perceivers can extract specific GUI modality signals from the visual clues of the visual backbone, we analyze the distribution discrimination in feature space. Specifically, we use t-SNE (t-Distributed Stochastic Neighbor Embedding) to visualize the GUI modality signals generated by different perceivers on images of downstream tasks, and the results are shown in \cref{visulation_tSNE}. It can be observed that the feature distributions are clearly distinguished into three groups, demonstrating that our method can extract different GUI modality information from the visual clues effectively.

\section{More Comparisons for Grounding Results}
Given that grounding ability serves as the foundation for MLLMs to attain more precise GUI understanding~\cite{seeclick}, in this section, we extend the evaluation metrics (Acc@IoU = 0.1) on the RefExp~\cite{refexp} benchmark. Specifically, we introduce Acc@IoU=0.3, Acc@IoU=0.5, Acc@IoU=0.7 and the Center Point Accuracy (Acc@CP) metrics to further assess the localization capabilities of diverse MLLMs. A larger IoU value (Acc@IoU=0.5/0.7) can quantify the degree of fit of the bounding box generated by the MLLM, and Acc@CP can reflect the model's ability to accurately click on the target area according to the instruction.
The formula of Acc@CP is defined as follows:
\begin{equation} 
Acc@CP = \frac{\sum_{i=1}^{n}\mathbb{I}(pred_i, gt_i)}{n} \times 100\%,
\end{equation}
where $\mathbb{I}(pred, gt)$ means an indicator function, which is used to calculate whether the center point of the predicted coordinates $pred$ is located inside $gt$.
\begin{figure*}
  \centering
  \includegraphics[width=1\linewidth]{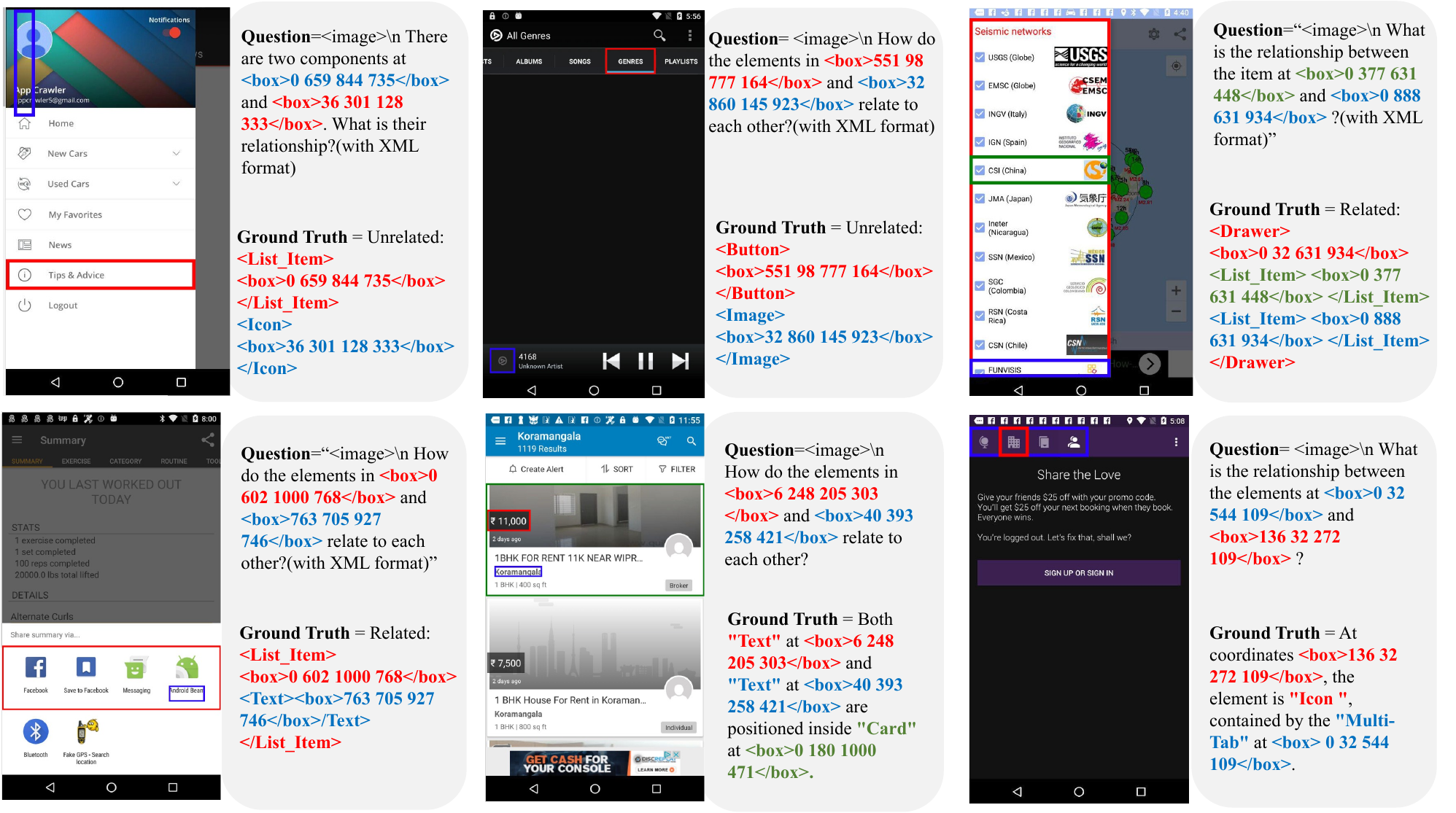}
  \caption{Examples of our SRP data.}
  \label{fig:SRP_data_samples}
\end{figure*}

\begin{table}[ht]\LARGE
\centering
\renewcommand{\arraystretch}{1.2}
\resizebox{\linewidth}{!}{
    \begin{tabular}{cccccc}
     \toprule
     & \textit{\textbf{IoU=0.1}}&\textit{\textbf{IoU=0.3}}&\textit{\textbf{IoU=0.5}}&\textit{\textbf{IoU=0.7}}&\textit{\textbf{Acc@CP}}  \\
    \toprule
    Qwen-VL~\cite{Qwen-VL} &36.3&25.3&16.3 & 9.2 & 59.3\\
    MiniCPM-V 2.6~\cite{yao2024minicpm} &48.5 &26.2&11.0& 2.5 & 66.5\\
    Qwen2-VL~\cite{Qwen2VL} &47.6 &36.2&27.7& 12.2 & \underline{86.5}\\
    Llama 3.2-V~\cite{Llama3}&51.3 &29.9&17.3& 9.6 &63.0 \\
    CogAgent~\cite{cogagent} &\underline{73.3} & \underline{68.0}&\underline{58.8}&\textbf{46.2}& 83.9\\
    InternVL2~\cite{chen2024far} &71.7 &52.9&35.7& 17.9& 74.9\\
    \model(ours) &\textbf{83.0} &\textbf{74.3}&\textbf{60.0 }& \underline{41.2} & \textbf{87.4}\\
    \bottomrule
    \end{tabular}
}
\caption{Evaluation of baseline MLLMs on RefExp~\cite{refexp} benchmark using different metrics. IoU=0.1/0.3/0.5/0.7 are shorthand for Acc@IoU=0.1/0.3/0.5/0.7 respectively.}
\label{ground_compare}
\end{table}
As shown in \cref{ground_compare}, although our \model(8B) achieves the second best result compared to CogAgent(18B)~\cite{cogagent} at the Acc@IoU=0.7 metric, still shows advanced performance overall.

 
\section{Spatial Relationship Prediction Examples} \label{SRP_samples}
To strengthen the pure visual MLLMs in perceiving the spatial relationship among elements on the screen, we introduce the Spatial Perceiver and SRP training tasks for explicit modeling of the spatial relationship (refer to \cref{sec:model_architecture} and~\ref{sec: multi_stage_training}). In this part, we display more SRP data samples (see \cref{fig:SRP_data_samples}). The SRP dataset is constructed using the VH json files corresponding to the images in the public dataset~\cite{rico_semantic}.

\begin{figure*}[ht]
  \centering
  \includegraphics[width=1\linewidth]{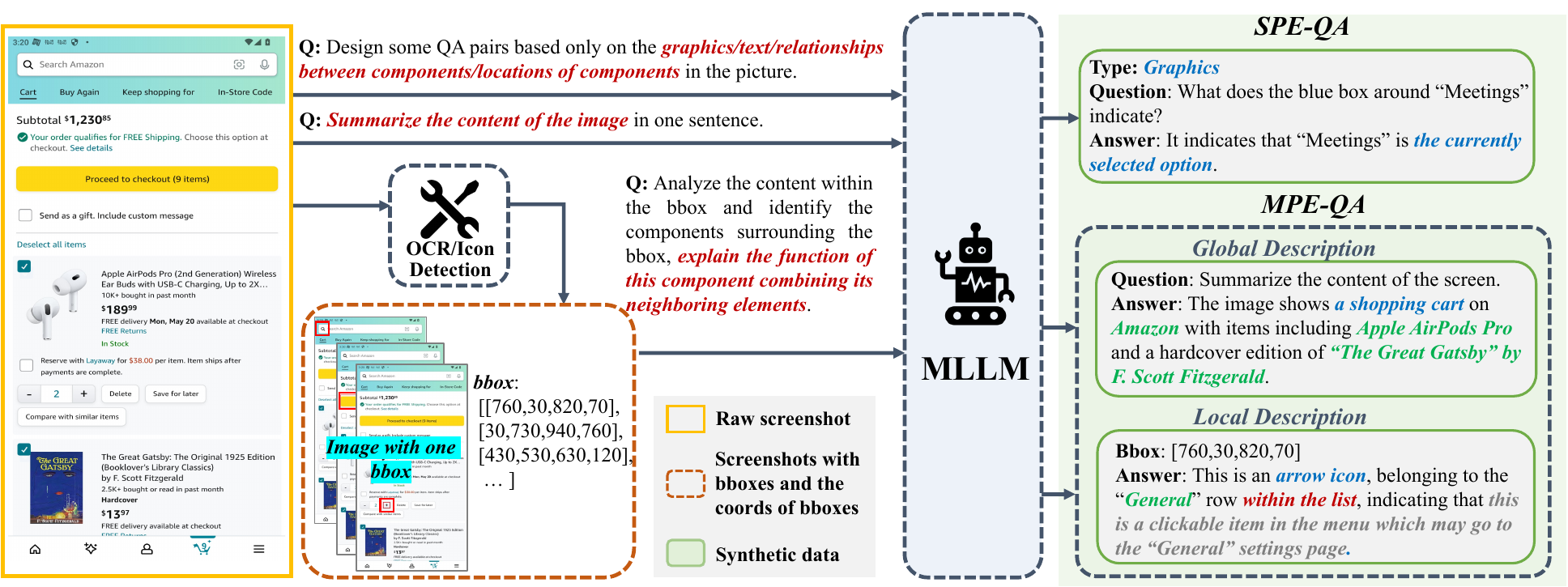}
  \caption{The pipeline for synthetic data generation. 
  We categorize the data into: SPE-QA (Single Perceiver Enhanced Question Answering) and MPE-QA (Multi-Perceiver Enhanced Question Answering). 
  }
  \label{fig:data_gen}
\end{figure*}
\section{Prompts in Automated Pipeline}\label{appendex_pipeline}
In this section, we present the prompts fed to Qwen2-VL (72B)~\cite{Qwen2VL} for generating Single Perceiver Enhanced Question Answering (SPE-QA) and Multi-Perceiver Enhanced Question Answering (MPE-QA) data, as introduced in \cref{sec:synthetic_data_generation}. The framework of the data synthesis pipeline is shown in \cref{fig:data_gen}. 

\subsection{SPE-QA}

\begin{tcolorbox}[colback=blue!5, title = {The prompt for \textbf{SPE-QA}}]
\textit{\tt Design some QA pairs based only on the icons in the picture, only on the text in the picture, only on some relationships between components and only on locations of components (such as the return icon is in the upper left corner of the screen.), and give questions and correct answers.}

\textit{\tt Please format the data as JSON format such as {'question': ..., 'type': 'text' or 'icon' or 'relationship' or 'location', 'answer': ...}.}
\end{tcolorbox}

\subsection{MPE-QA}

\begin{tcolorbox}[colback=blue!5, title = {The prompt for \textbf{Global Description}}]
\textit{\tt Generate a summary of the screen in one sentence. Do not focus on specifically naming the various UI elements, but instead, focus on the content.}
\end{tcolorbox}

\begin{tcolorbox}[colback=blue!5,title = {The prompt for \textbf{Local Description}},breakable]
\textit{\tt Describe this image. You will receive a screenshot 
of a GUI that includes a bounding box (bbox) with specified coordinates. Your task is to analyze the content within the bbox and identify the component to which it belongs by looking for surrounding component boundaries. Please provide a detailed description that includes the following:}

\begin{quote}
    \textit{\tt 1.Identify the content inside the bbox (text or graphic element).} 
    
    \textit{\tt 2.Look for the component boundary surrounding the bbox and describe the overall component it belongs to.}
    
    \textit{\tt 3.Explain the function of this component and any other relevant elements it contains.} 
    
    \textit{\tt 4.If there are no surrounding component boundaries, state that there are no related components nearby.}
\end{quote}

\textit{\tt Output Example (response with just one sentence):}
\begin{quote}
    \textit{\tt "This is an icon of a house, belonging to a button component that describes the home page; it also includes another house icon as part of this component."}
    
    \textit{\tt "This is an arrow icon, belonging to the 'General' row within the list, indicating that this is a clickable item in the menu which may go to the 'General' page."}
    
    \textit{\tt "This is a standalone button labeled 'Submit', and there are no related components nearby."}
\end{quote}

\textit{\tt Now the coordinate of bbox I'd like you to analyze is \textbf{[bbox]}}
\end{tcolorbox}

\section{More Qualitative Analysis} \label{more_qualitative}
 In this section, we show more qualitative results of our \model with other MLLMs on downstream tasks.

\begin{figure}
    \centering
    \includegraphics[width=1\linewidth]{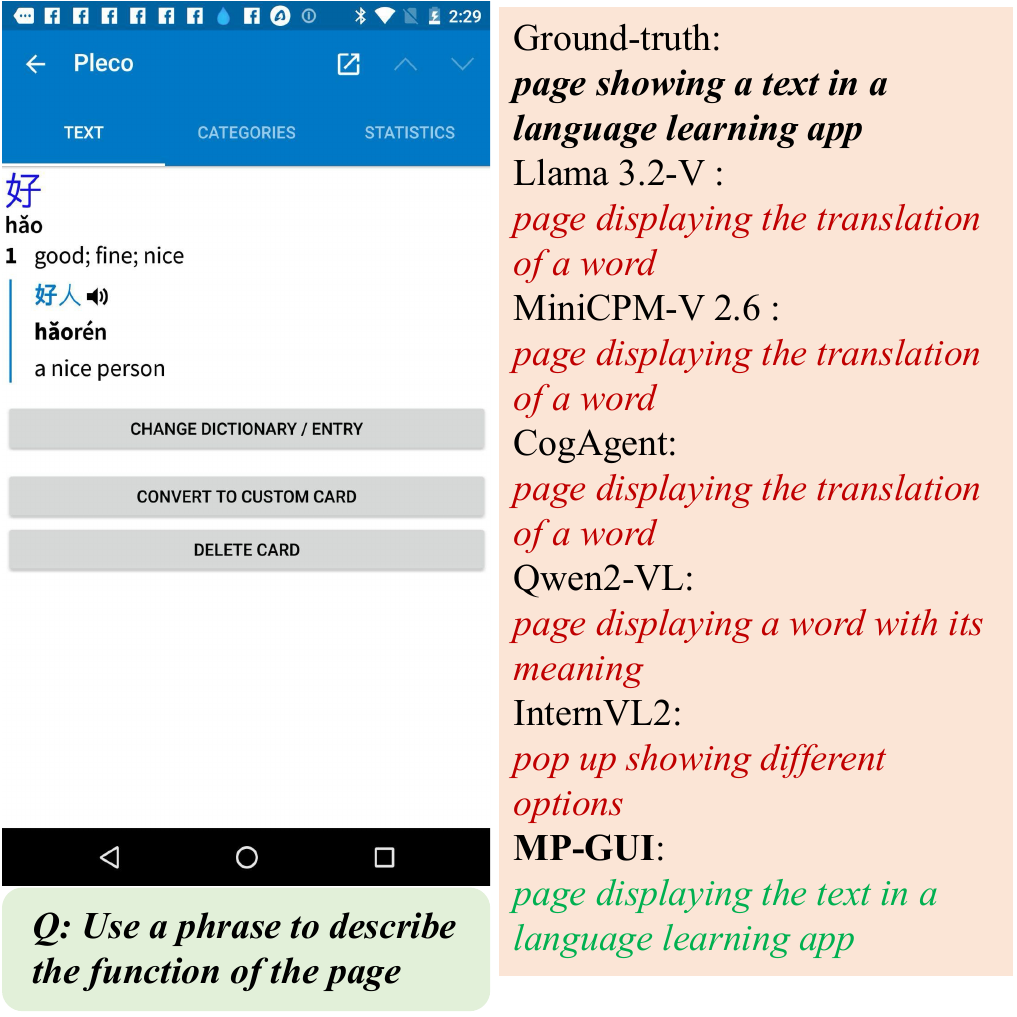}
    \caption{A comparison on Screen2Words~\cite{screen2words}.}
    \label{fig:s2w}
\end{figure}

\noindent \textbf{Screen2Words.} As shown in \cref{fig:s2w}, \model is capable of taking into account the overall layout and determining that the page belongs to the language learning app. In contrast, all other methods are distracted by the sizable translation portion in the middle of the screen.

\begin{figure*}[ht]
    \centering
    \includegraphics[width=0.9\linewidth]{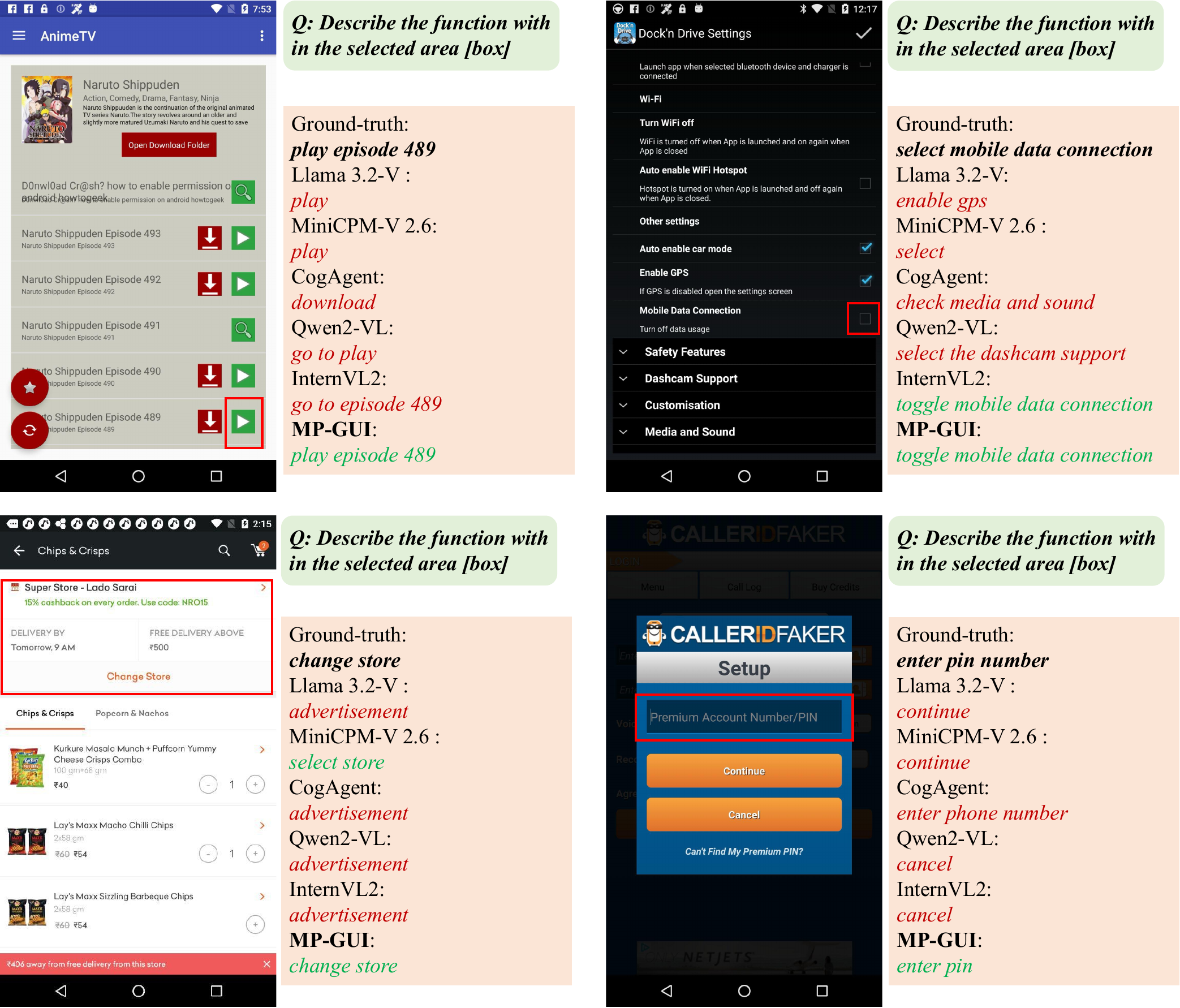}
    \caption{Comparisons on Widget Captioning~\cite{widgetcaption}.}
    \label{fig:wc}
\end{figure*}

\noindent \textbf{Widget Captioning.} As depicted in \cref{fig:wc}, under the guidance of the novel Local Description task (see \cref{sec:synthetic_data_generation}), our \model is more inclined to summarize the graphics by combining the spatial context information. In the first example, \model can summarize the high-level function of \textit{"play episode 489"} by combining the text on the left of the button, instead of only focusing on the graphical element \textit{"play"}. Meanwhile, our method is also capable of differentiating the core content within the target area, as demonstrated in the third example. Furthermore, due to the excellent grounding ability, \model is able to precisely comprehend the coordinates in the input question and provide accurate answers, rather than misidentifying the location as \textit{"dashcam support"} (in Example 2) or \textit{"continue"} (in Example 4).

\begin{figure*}
    \centering
    \includegraphics[width=0.9\linewidth]{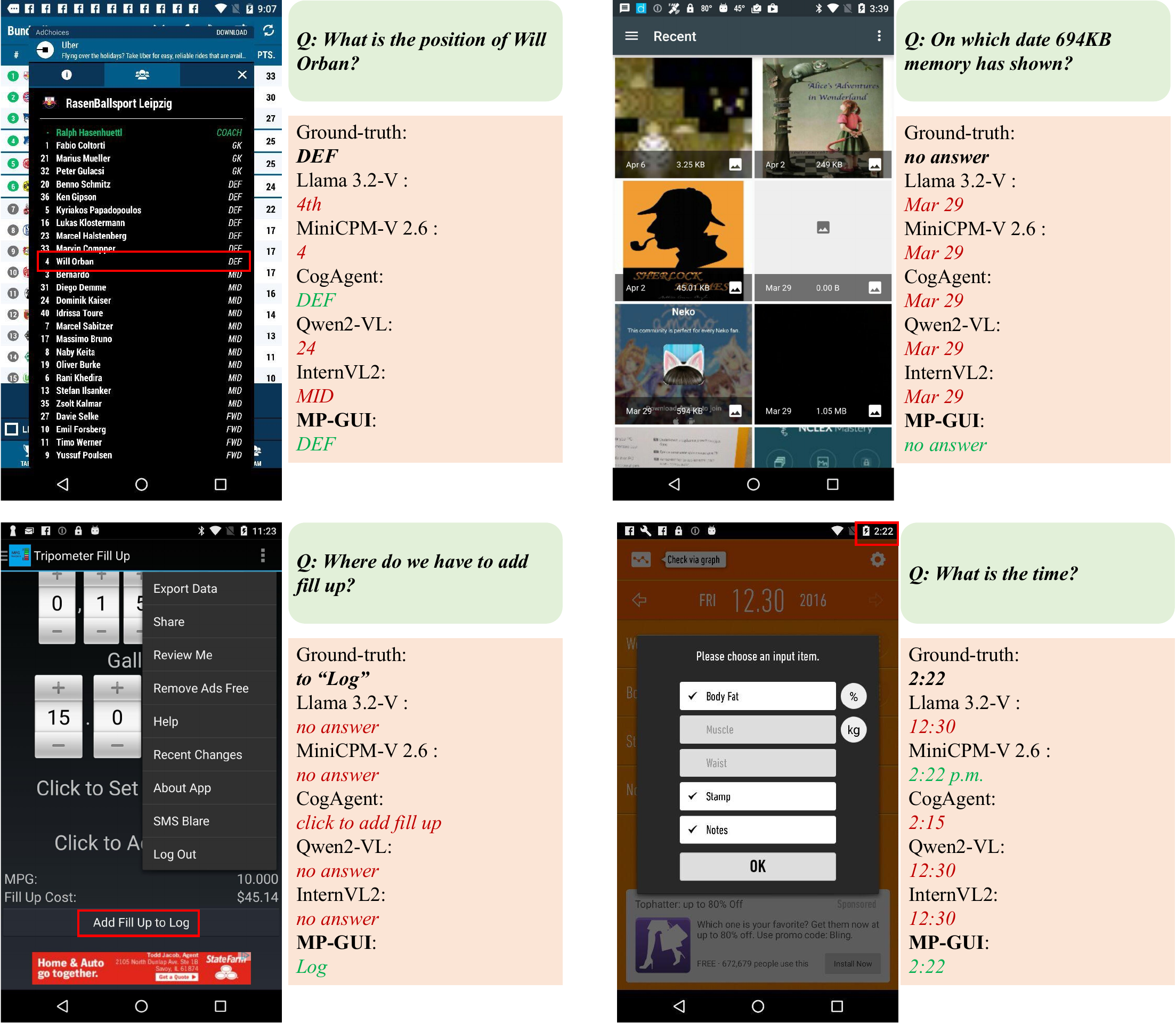}
    \caption{Comparisons on ScreenQA Short~\cite{screenai}.}
    \label{fig:qas}
\end{figure*}

\noindent \textbf{ScreenQA Short.} In the scenarios presented in \cref{fig:qas}, we observe that \model exhibits favorable OCR and comprehension capabilities. The Graphical Perceiver boosts the model's capacity to center on smaller areas. In contrast, Llama3.2-V(11B)~\cite{Llama3}, Qwen2-VL(7B)~\cite{Qwen2VL}, and InternVL2(8B)~\cite{chen2024far} are influenced by the sizable '\textit{12:30}' in the middle of the screen (as seen in Example 4). It is noteworthy that even when the question is unanswerable, as shown in the second example, our method still functions robustly.

\begin{figure*}
    \centering
    \includegraphics[width=0.9\linewidth]{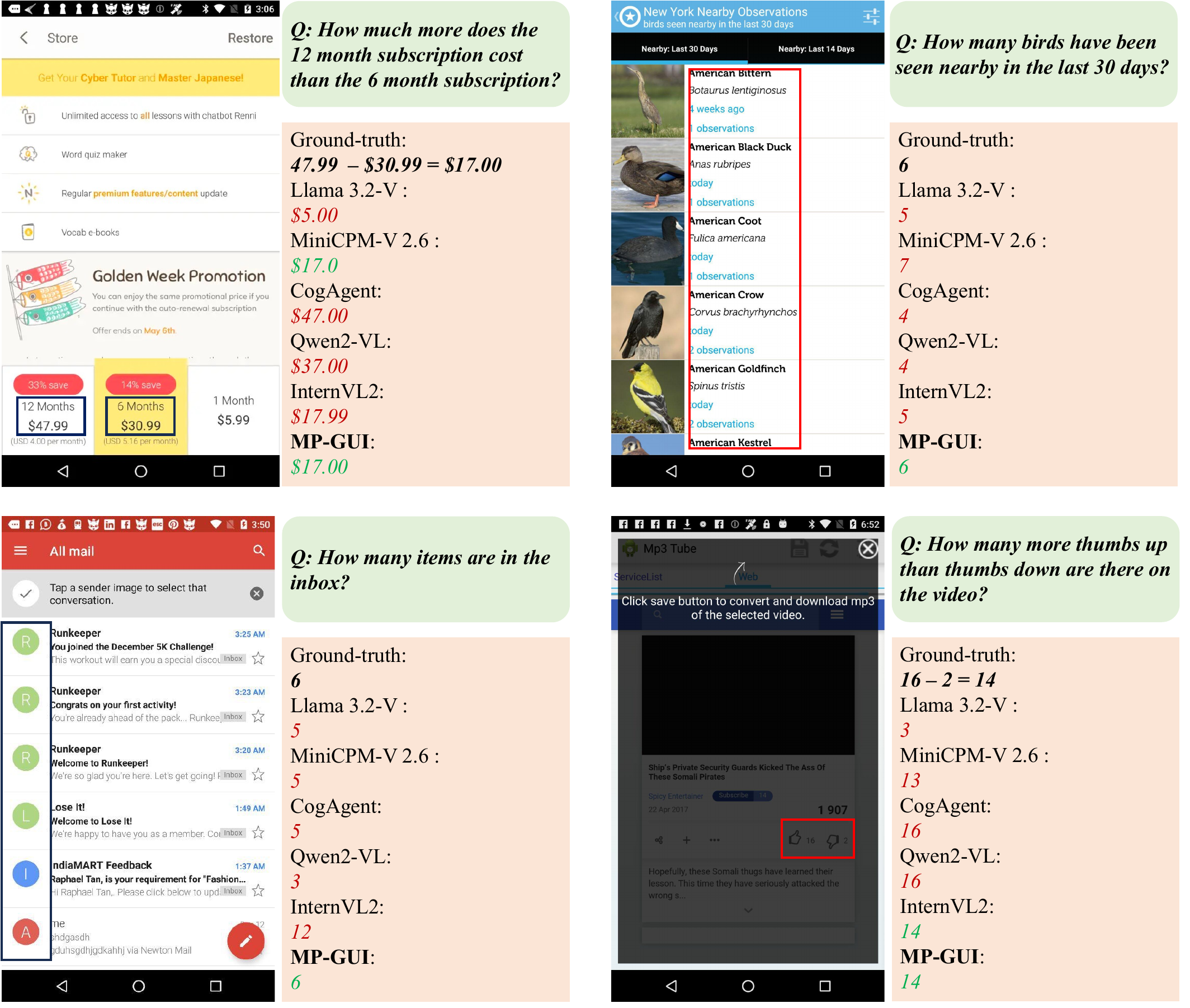}
    \caption{Comparisons on Complex ScreenQA~\cite{screenai}.}
    \label{fig:cqa}
\end{figure*}

\begin{figure*}
    \centering
    \includegraphics[width=0.9\linewidth]{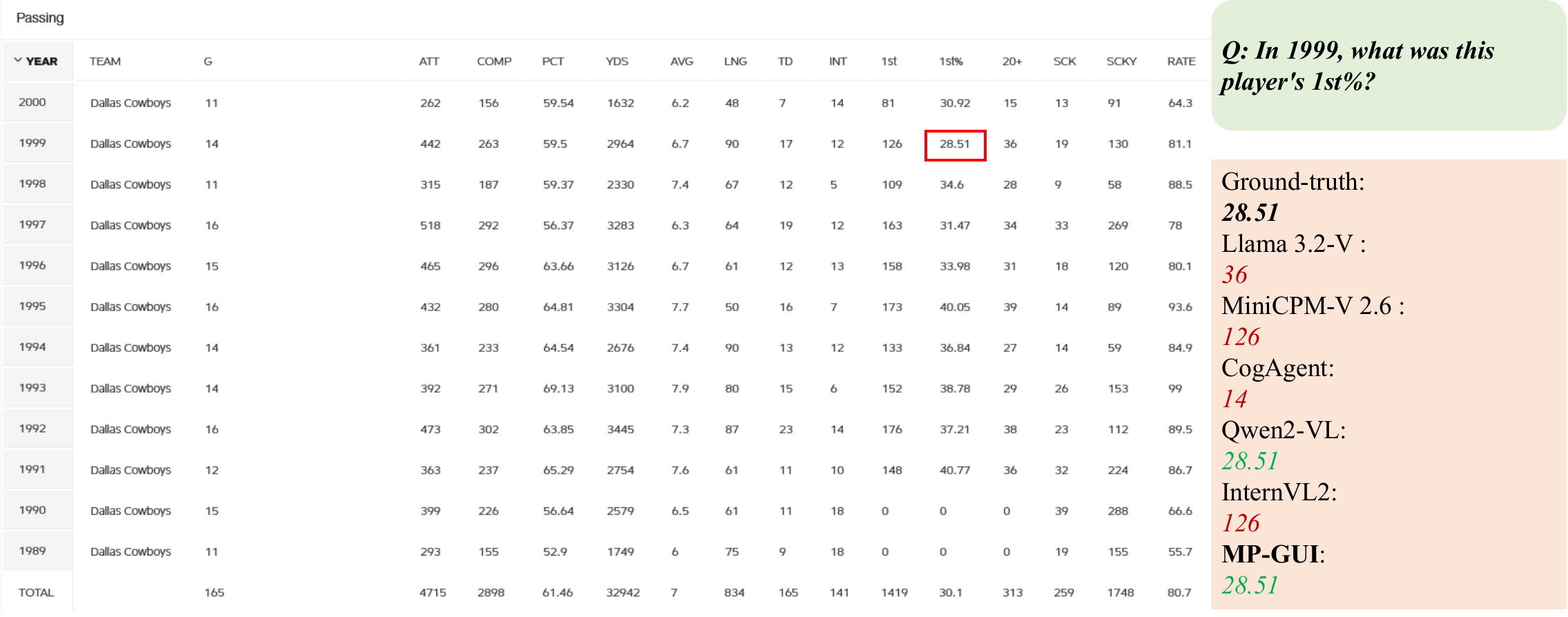}
    \hfill
    \includegraphics[width=0.5\linewidth]{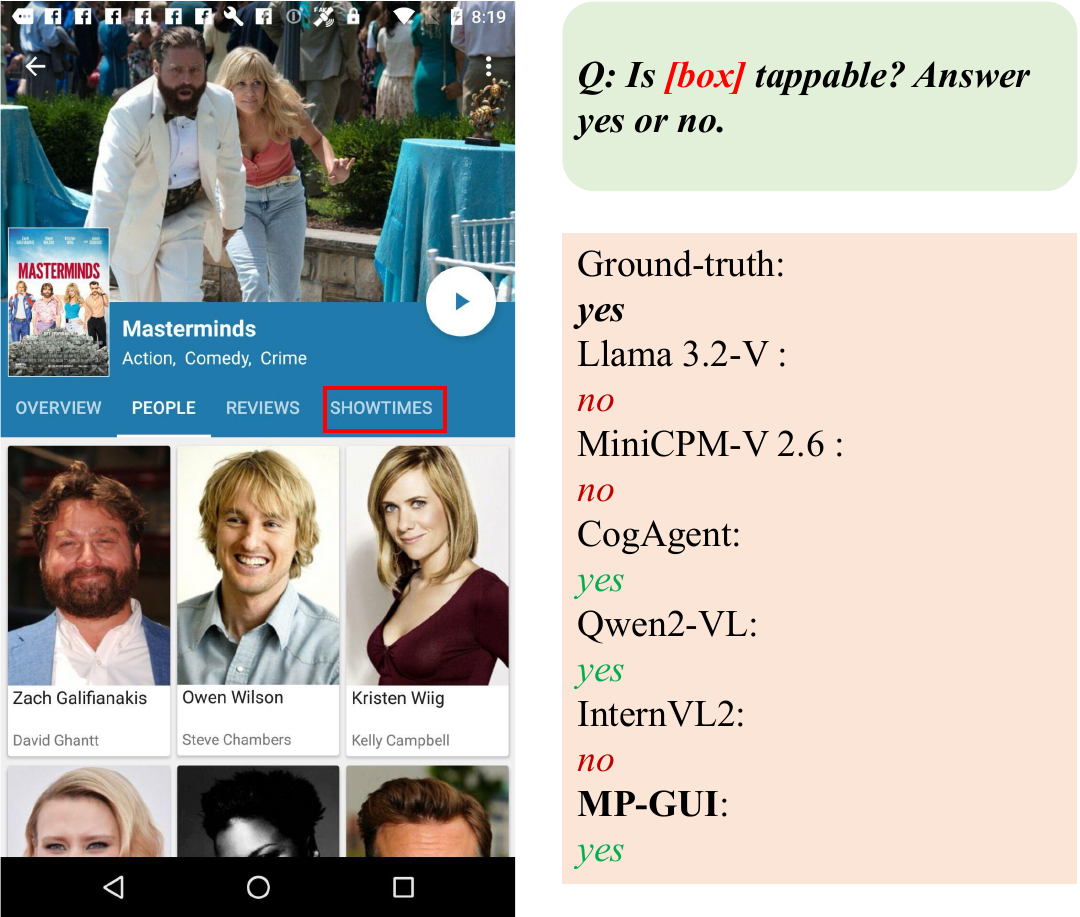}
    \caption{More qualitative results.}
    \label{fig:more}
\end{figure*}

\noindent \textbf{Complex ScreenQA.} The Spatial Perceiver enhances the awareness of spatial relationships between GUI elements on the screen. Compared with other MLLMs, our \model has advantages in difference calculation (as shown in Examples 1 and 4) and quantity counting (as shown in Examples 2 and 3) in \cref{fig:cqa}. More qualitative results of our \model are shown in \cref{fig:more}. 

       \end{document}